%% file: main.tex
\documentclass[round,authoryear,numbers]{ifm-tr}
\input{math_commands.tex}

\usepackage{multirow}
\usepackage{booktabs}
\usepackage{tikz}
\usepackage{amsmath, amssymb}
\usepackage{xcolor}
\usepackage{fontawesome5}
\newcommand{\gcirc}{\tikz\draw[fill=green!70!black,draw=green!70!black] circle (0.12cm);}
\newcommand{\rcirc}{\tikz\draw[fill=red!70!black,draw=red!70!black] circle (0.12cm);}

\newcommand{\mixcirc}{\rcirc\gcirc}

\usepackage{eccvabbrv}
\usepackage{times}
\usepackage{epsfig}
\usepackage{makecell}
\usepackage{graphicx}
\usepackage{hhline}
\usepackage{amsmath}
\usepackage{amssymb}
\usepackage{booktabs}
\usepackage{multirow}
\usepackage{algorithm}
\usepackage{algorithmic}
\usepackage{xcolor}
\usepackage{wrapfig} 
\usepackage{pifont} 
\usepackage{multicol}

\usepackage{thmtools}          
\usepackage{enumitem}          
\usepackage{siunitx}           
\usepackage{tikz}              
\usepackage{booktabs}          
\usepackage{subcaption}  
\usepackage{comment}
\usepackage{xcolor,colortbl}

\definecolor{lightred}{rgb}{1,0.8,0.8}
\definecolor{lightgreen}{rgb}{0.8,1,0.8}
\definecolor{lightblue}{rgb}{0.88,0.96,1}
\definecolor{lightgray}{rgb}{0.9,0.9,0.9}

\definecolor{darkgreen}{RGB}{0,100,0}

\usepackage{bbm}
\usepackage{float}

\usepackage{url}            

\usepackage{xurl}          
\usepackage{booktabs}       
\usepackage{amsfonts}       
\usepackage{nicefrac}       
\usepackage{microtype}      
\usepackage{graphicx}
\usepackage{subcaption} 
\usepackage{xspace}
\usepackage[frozencache,cachedir=.]{minted}
\usepackage{longtable}

\usepackage{nicematrix}
\usepackage{multirow}
\usepackage{soul}  
\usepackage{wrapfig}
\usepackage{enumitem}

\usepackage{lipsum}

\usepackage{colortbl}
\usepackage{ulem}
\usepackage{floatflt}
\usepackage{xspace}

\definecolor{cellHighlight}{HTML}{dbefff}

\newcommand\rurl[1]{%
    \href{https://#1}{\nolinkurl{#1}}%
}

\usepackage{xparse}

\title{SafeDiffusion-R1: Online Reward Steering for Safe \\ 
\vspace{3pt}
 Diffusion Post-Training}

\author[]{\textbf{Komal Kumar$^{1}$,\, Ankan Deria$^1$,\, Abhishek Basu$^1$,\, Fahad Shamshad$^1$,\, \\
Hisham Cholakkal$^1$,\, Karthik Nandakumar$^{1,2}$ }}\authorsep{}

\affiliation[]{$^1$Mohamed bin Zayed University of Artificial Intelligence (MBZUAI), UAE\\
$^2$Michigan State University (MSU), USA\\
\small
\begin{tabular}{cc}
{\fontsize{10}{10}\selectfont\faGithub}~\textbf{GitHub:}~\href{https://github.com/MAXNORM8650/SafeDiffusion-R1}{\texttt{\textcolor{teal}{https://github.com/MAXNORM8650/SafeDiffusion-R1}}}
\end{tabular} \\[2pt]
\small
{\fontsize{10}{10}\selectfont\faGlobe}~\textbf{Website:}~\href{https://maxnorm8650.github.io/SafeDiffusion-R1/}{\texttt{\textcolor{teal}{maxnorm8650.github.io/SafeDiffusion-R1/}}}\\
{\tt\small \{komal.kumar\}@mbzuai.ac.ae} \\[4pt]
}

\abstract{
Diffusion models have been widely studied for removing unsafe content learned during pre-training. Existing methods require expensive supervised data, either unsafe-text paired with safe-image groundtruth or negative/positive image pairs, making them impractical to scale. Furthermore, offline reinforcement learning and supervised fine-tuning approaches that generate synthetic data offline suffer from catastrophic forgetting, degrading generation quality. We propose a novel online reinforcement learning framework that addresses both data scarcity and model degradation through post-training with Group Relative Policy Optimization (GRPO) on both negative and positive text prompts. To eliminate the need for fine-tuning specialized safe/unsafe reward models, we introduce a \textit{steering reward mechanism} that exploits an inherent property of CLIP embeddings: steering text representations toward positive safety directions and away from negative ones in the embedding space. Our online-policy approach enables the model to learn from diverse prompts, including explicit unsafe content, without catastrophic forgetting. Extensive experiments demonstrate that our method reduces inappropriate content to 18.07\% (vs. 48.9\% for SD v1.4) and nudity detections to 15 (vs. 646 baseline) while improving compositional generation quality from 42.08\% to 47.83\% on GenEval. Remarkably, these safety gains generalize to out-of-domain unsafe prompts across seven harm categories, achieving state-of-the-art performance without supervised paired data or reward tuning.}

\begin{document}
\maketitle
\input{sections/2_introduction}
\input{sections/3_related_work}

\input{sections/4_methodology}

\let\oldvspace\vspace
\RenewDocumentCommand{\vspace}{s m}{%
}
\input{sections/5_experiments}
\input{sections/6_conclusion}
\bibliographystyle{plainnat}
\bibliography{main}
\clearpage
\input{sections/supplymentry}
\end{document}

%% file: math_commands.tex
\usepackage{amsmath,amsfonts,bm}

\def\eqref#1{equation~\ref{#1}}

\def\1{\bm{1}}

\DeclareMathAlphabet{\mathsfit}{\encodingdefault}{\sfdefault}{m}{sl}
\SetMathAlphabet{\mathsfit}{bold}{\encodingdefault}{\sfdefault}{bx}{n}



%% file: sections/2_introduction.tex
\section{Introduction}
\label{sec:intro}

The rapid advancement of text-to-image (T2I) diffusion models~\citep{rombach2022high, ramesh2022hierarchical, saharia2022photorealistic,ho2020denoising} has democratized high-quality visual content generation. 
Trained on large-scale web data, these models learn rich multimodal representations that enable controllable generation across a wide range of concepts. However, this broad representational capacity also leads them to internalize unsafe and explicit associations from the data, which can be triggered by explicit or inappropriate textual prompts. The public availability of T2I models such as Stable Diffusion (SD)~\citep{rombach2022high} further amplifies these risks, raising significant safety concerns that demand effective mitigation strategies.
Existing safety interventions for T2I diffusion models generally fall into three categories: dataset filtering before training, output filtering, and post-training model modification. 
Dataset filtering~\citep{carlini2022privacy} removes unsafe content from the training corpus before training diffusion model but is computationally expensive at scale and difficult to extend to newly emerging or long-tail concepts. 
\begin{center}
    \centering
    \includegraphics[width=1.0\textwidth]{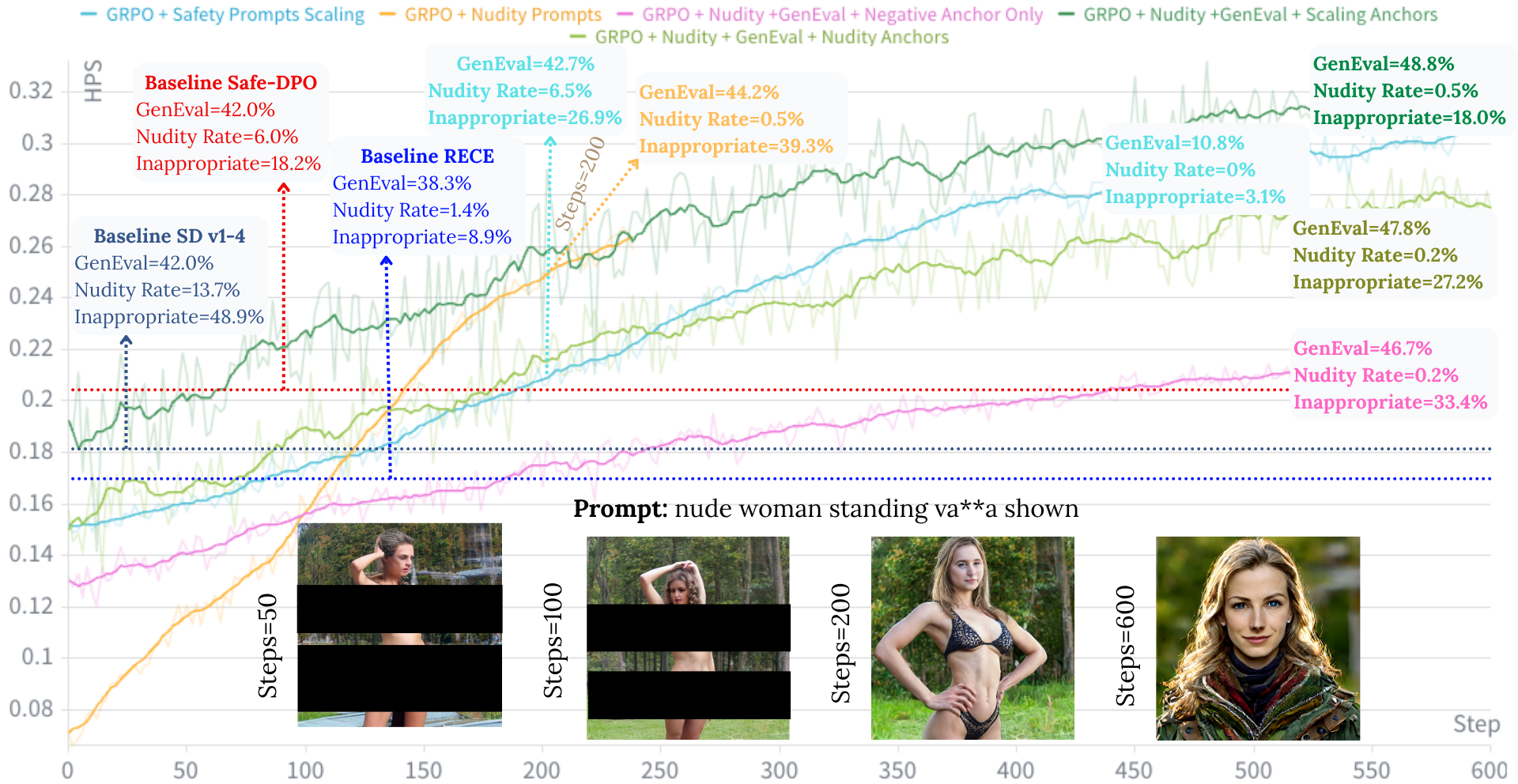}
\captionof{figure}{\small  
Effect of post-training reward design on safety--utility trade-off. Each curve tracks HPSv2~\citep{wu2023human} over GRPO~\citep{shao2024deepseekmath,Xue2025DanceGRPOUG} training steps; annotations report GenEval, Nudity Rate, and Inappropriate rate at key
checkpoints. Horizontal lines denote static baselines (SD v1.4, Safe-DPO, RECE). \textcolor{cyan}{\textit{Safety Prompt Scaling}} uses diverse safety prompts
(harassment, shocking, nudity, etc.) from SafeDPO~\citep{liu2025alignguard}.
\textcolor{orange}{\textit{Nudity Prompts}} converges rapidly due to limited data.
The remaining variants fix the training data to GenEval-style + negative prompts,
varying only the anchor design:
\textcolor{magenta}{\textit{Negative Anchor Only}} uses an empty prompt ``.'',
\textcolor{darkgreen}{\textit{Scaling Anchors}} uses diverse safety anchors generated via ChatGPT,
and \textcolor{olive}{\textit{Nudity Anchors}} uses nudity-specific anchors. Notably, all variants except \textit{\textcolor{cyan}{Safety Prompt Scaling}} are trained exclusively on nudity prompts, yet achieve broad inappropriate content reduction, demonstrating strong OOD generalization.
}
    \label{fig:intro_fig}
\end{center}%

Output filtering~\citep{schramowski2023safe} suppresses harmful generations at inference time, yet leaves the underlying generative distribution unchanged and offers limited robustness under direct model access. As a result, post-training modification has emerged as the most practical strategy~\citep{gandikota2023erasing}, directly adjusting pre-trained models to suppress unsafe concepts without retraining from scratch and remaining compatible with publicly released systems such as Stable Diffusion.

Among these post-training methods~\citep{kumar2025llm}, supervised fine-tuning~\citep{kumar2025deft} and offline reinforcement learning~\citep{cho2024_456} have become the dominant paradigms for safety alignment. Supervised fine-tuning relies on curated safe/unsafe examples~\citep{schramowski2023safe, qu2023unsafe}, while offline reinforcement learning optimizes the model against a fixed reward signal using pre-generated data~\citep{black2023training, clark2023directly}. However, from the perspective of concept unlearning, both approaches are inherently limited, as neither paradigm adapts its training signal to the model's current generative behavior: supervised fine-tuning optimizes on fixed examples regardless of what the model currently produces, and offline RL optimizes against rewards computed on pre-generated data rather than on-policy samples. This static supervision is insufficient to track and suppress unsafe content that emerges as the model evolves during training. Ideally, concept unlearning should be formulated as an online process, in which the model continuously generates samples during training, receives feedback on its current outputs, and progressively reduces the discrepancy between its realized generations and the desired safety constraints. Furthermore, offline reinforcement learning methods often require training or fine-tuning specialized reward models to classify images as safe or unsafe, introducing additional computational overhead.
\begin{table*}[t]
\centering
\small
\caption{\textbf{Comparison of safety methods for text-to-image diffusion models.} Our approach eliminates the need for supervised paired data, prevents catastrophic forgetting through online-policy training, requires no reward model fine-tuning, and achieves superior generalization to out-of-domain unsafe prompts.}
\label{tab:comparison}
\resizebox{\textwidth}{!}{
\begin{tabular}{l|cccccc}
\toprule
\textbf{Method} &
\begin{tabular}{@{}c@{}}\textbf{Supervised}\\\textbf{Data}\end{tabular} &
\begin{tabular}{@{}c@{}}\textbf{Training}\\\textbf{Policy}\end{tabular} &
\begin{tabular}{@{}c@{}}\textbf{Catastrophic}\\\textbf{Forgetting}\end{tabular} &
\begin{tabular}{@{}c@{}}\textbf{Reward Model}\\\textbf{Fine-tuning}\end{tabular} &
\begin{tabular}{@{}c@{}}\textbf{Reasoning}\\\textbf{Capability}\end{tabular} &
\begin{tabular}{@{}c@{}}\textbf{OOD}\\\textbf{Generalization}\end{tabular} \\
\midrule

Post-hoc Filtering~\citep{schramowski2023safe}
    & \gcirc \gcirc        
    & N/A
    & \gcirc \gcirc       
    & \rcirc \rcirc        
    & \rcirc \rcirc        
    & \mixcirc \\   

Concept Erasure~\citep{qu2023unsafe}
    & \rcirc \rcirc       
    & Offline
    & \rcirc \rcirc       
    & \gcirc \gcirc       
    & \rcirc \rcirc  
    & \rcirc \rcirc \\     

Supervised Fine-tuning~\citep{schramowski2023safe}
    & \rcirc \rcirc       
    & Offline
    & \rcirc \rcirc       
    & \rcirc \rcirc       
    & \rcirc \rcirc       
    & \rcirc \rcirc \\     

Prompt Filtering~\citep{lee2023aligning}
    & \gcirc \gcirc        
    & Online/Offline
    & \rcirc \rcirc        
    & \rcirc \rcirc       
    & \rcirc \rcirc       
    & \rcirc \rcirc \\     

Offline-RL (DDPO)~\citep{black2023training}
    & \rcirc \rcirc        
    & Offline
    & \rcirc \rcirc        
    & \rcirc  \rcirc       
    & \mixcirc      
    & \mixcirc \\   

DPO~\citep{clark2023directly}
    & \rcirc \rcirc       
    & Offline
    & \rcirc \rcirc        
    & \gcirc \gcirc       
    & \mixcirc      
    & \mixcirc \\   

SafetyDPO~\citep{kim2025safedpo,liu2025alignguard}
    &  \rcirc\gcirc             
    & Offline              
    &  \gcirc \gcirc            
    &  \rcirc\gcirc             
    & \mixcirc      
    & \rcirc \rcirc \\            

AttnSteering~\citep{gaintseva2025casteer}
    &  \gcirc \gcirc            
    &  N/A             
    &  \gcirc \gcirc            
    &     N/A          
    & \rcirc \rcirc        
    & \rcirc \rcirc \\            

\midrule
\textbf{Ours (GRPO + Steering)}
    & \gcirc \gcirc       
    & \textbf{Online}
    & \gcirc \gcirc       
    & \gcirc \gcirc       
    & \textbf{\gcirc \gcirc}        
    & \textbf{\gcirc \gcirc} \\ 

\bottomrule
\end{tabular}
}

\vspace{0.4em}
\noindent\small
\gcirc \gcirc~Favorable \quad \rcirc \rcirc~Unfavorable \quad \mixcirc~Partial

\vspace{-10 pt}
\end{table*}

To address these limitations, we propose \textbf{SafeDiffusion-R1}, an online reinforcement learning framework for safe text-to-image generation that avoids reliance on static datasets or additional reward-model fine-tuning. Our approach consists of two key components. \textbf{First}, we adopt Group Relative Policy Optimization (GRPO)~\citep{shao2024deepseekmath} as an online policy optimization algorithm, in which the model continuously generates images from both benign and unsafe prompts and receives feedback on its current outputs. By directly coupling safety optimization with the model’s evolving sampling distribution, this on-policy formulation mitigates distribution mismatch and enables the model to preserve its general generative capabilities while progressively unlearning unsafe concepts. \textbf{Second}, we introduce a geometry-aware steering reward that eliminates the need for a separately trained safe/unsafe classifier. Leveraging a structural property of CLIP~\citep{radford2021learning}, we represent safety as a direction in text embedding space, estimated from a small set of contrastive safe and unsafe descriptions. During training, embeddings of unsafe prompts are steered toward this safe direction prior to reward computation, reshaping the optimization signal without explicitly rewarding unsafe image generation. The steering operates purely through embedding manipulation and requires no additional model training. Tab.~\ref{tab:comparison} compares our method with existing safety approaches, while Fig.~\ref{fig:intro_fig} illustrates the post-training capabilities enabled by our approach. We name our method \textbf{SafeDiffusion-R1} to reflect its dual objective: improving safety in diffusion post-training while enhancing reward-guided reasoning for safer and more reliable image generation. \textit{Together, our online GRPO training with steering rewards eliminates the need for supervised safety datasets and reward-model fine-tuning, mitigates catastrophic forgetting through on-policy optimization, and improves generalization to out-of-domain unsafe prompts}.

Our contributions can be summarized as follows:

\begin{enumerate}
    \item We formulate safety alignment for text-to-image diffusion models as an \textbf{online policy optimization} problem and introduce a GRPO-based training framework that couples safety learning with the model’s evolving generative distribution.

    \item We propose a \textbf{geometry-aware steering reward} that represents safety as a direction in CLIP embedding space, enabling concept suppression without training dedicated safe/unsafe reward models.


    \item We conduct extensive empirical analysis demonstrating that our online policy optimization framework consistently outperforms supervised fine-tuning and offline alignment methods on standard safety benchmarks, while preserving generation quality on benign concepts.
\end{enumerate}


The remainder of this paper is organized as follows. 
Section~\ref{sec:related} reviews related work. 
Section~\ref{sec:method} presents our methodology, including the steering reward formulation (Section~\ref{sec:steering}) and the GRPO framework (Section~\ref{sec:grpo}). 
Section~\ref{sec:experiments} describes the experimental setup. 
Section~\ref{sec:results} reports the results and analysis. 
Section~\ref{sec:conclusion} concludes the paper.

%% file: sections/3_related_work.tex
\section{Related Work}
\label{sec:related}

\textbf{Harmful Concept Erasing from Diffusion Models.} T2I diffusion models can be misused to generate unsafe content, including sexually explicit imagery, harassment, and depictions of illegal activities~\citep{liu2024machine,huang2025survey}. Early systems used post-hoc NSFW filters, which only screen outputs, leave the model unchanged, and can be bypassed with direct access~\citep{rando2022red}. More principled methods modify model parameters to remove harmful concepts without full retraining. Safe Latent Diffusion (SLD)~\citep{schramowski2023safe} applies inference-time guidance to steer denoising away from unsafe semantic directions. Post-training parameter editing methods directly alter model weights to erase unsafe associations. ESD~\citep{gandikota2023erasing} fine-tunes UNet weights to suppress targeted concepts, with ESD-x targeting cross-attention layers and ESD-u modifying unconditional score predictions. UCE~\citep{DBLP:conf/wacv/GandikotaOBMB24} and Ablating Cross-Attention (CA)~\citep{DBLP:conf/iccv/KumariZWS0Z23} perform structured weight updates to localize suppression while preserving unrelated content. SA~\citep{DBLP:conf/nips/HengS23}, RECE~\citep{DBLP:conf/eccv/GongCWCJ24}, MACE~\citep{DBLP:conf/cvpr/LuWLLK24}, Receler~\citep{huang2023receler}, CPE~\citep{lee2024cpe}, STEREO~\citep{srivatsan2025stereo}, and SAeUron~\citep{cywinski2025saeuron} further refine these strategies through parameter-efficient, closed-form, or feature-level editing to better preserve benign semantics. Safe-DPO~\citep{liu2025alignguard} adapts direct preference optimization to diffusion safety, framing concept suppression as a preference alignment problem; however, its reliance on fixed preference datasets provides static supervision that is often insufficient to track and suppress unsafe content that emerges as the model evolves during training. 

\noindent \textbf{Reinforcement Learning for Diffusion Models.} Reinforcement learning has emerged as an effective paradigm for aligning generative models with objectives that are difficult to capture through supervised losses alone~\citep{ouyang2022training, bai2022training}. Extending RL to diffusion models is considerably more challenging than in autoregressive language models due to the multi-step denoising process, which involves long-horizon credit assignment across timesteps. DDPO~\citep{black2023training} adapts PPO~\citep{schulman2017proximal} to optimize diffusion trajectories using image-level rewards, while DPOK~\citep{fan2023dpok} introduces KL regularization to mitigate reward over-optimization. Clark et al.~\citep{clark2023directly} further extend Direct Preference Optimization to diffusion models, eliminating explicit reward-model training via pairwise preference learning. However, these approaches primarily target aesthetic quality or prompt alignment rather than safety. When safety is addressed, it is typically handled through dataset curation or prompt filtering: for example, training exclusively on curated safe prompts~\citep{lee2023aligning} or excluding NSFW content during reward-model training~\citep{xu2023imagereward}. As a result, the learned policy is not explicitly optimized for unsafe inputs and may generalize poorly.

Unlike offline methods~\citep{liu2025alignguard} that rely on fixed datasets, online policy optimization updates the model using its own current outputs. This setting introduces a key challenge: rewards for unsafe prompts typically exhibit higher magnitude and variance than those for benign prompts, causing standard PPO-style updates to overcorrect and degrade unrelated concepts. GRPO~\citep{shao2024deepseekmath} mitigates this instability by normalizing advantages within groups of generations from the same prompt, making updates depend on relative comparisons rather than absolute reward scale. This property is crucial for safety unlearning, where harmful concepts must be suppressed without globally shifting the model’s distribution. Moreover, many offline RL approaches require training or fine-tuning dedicated safe/unsafe reward models, introducing additional computational overhead. In contrast, we apply online GRPO-based optimization with a geometry-aware CLIP reward, enabling targeted concept suppression that generalizes beyond the unsafe prompts observed during training without separate reward-model training.

%% file: sections/4_methodology.tex
\section{Methodology}
\label{sec:method}

\begin{figure}[t]
    \centering
    \includegraphics[width=0.94\linewidth]{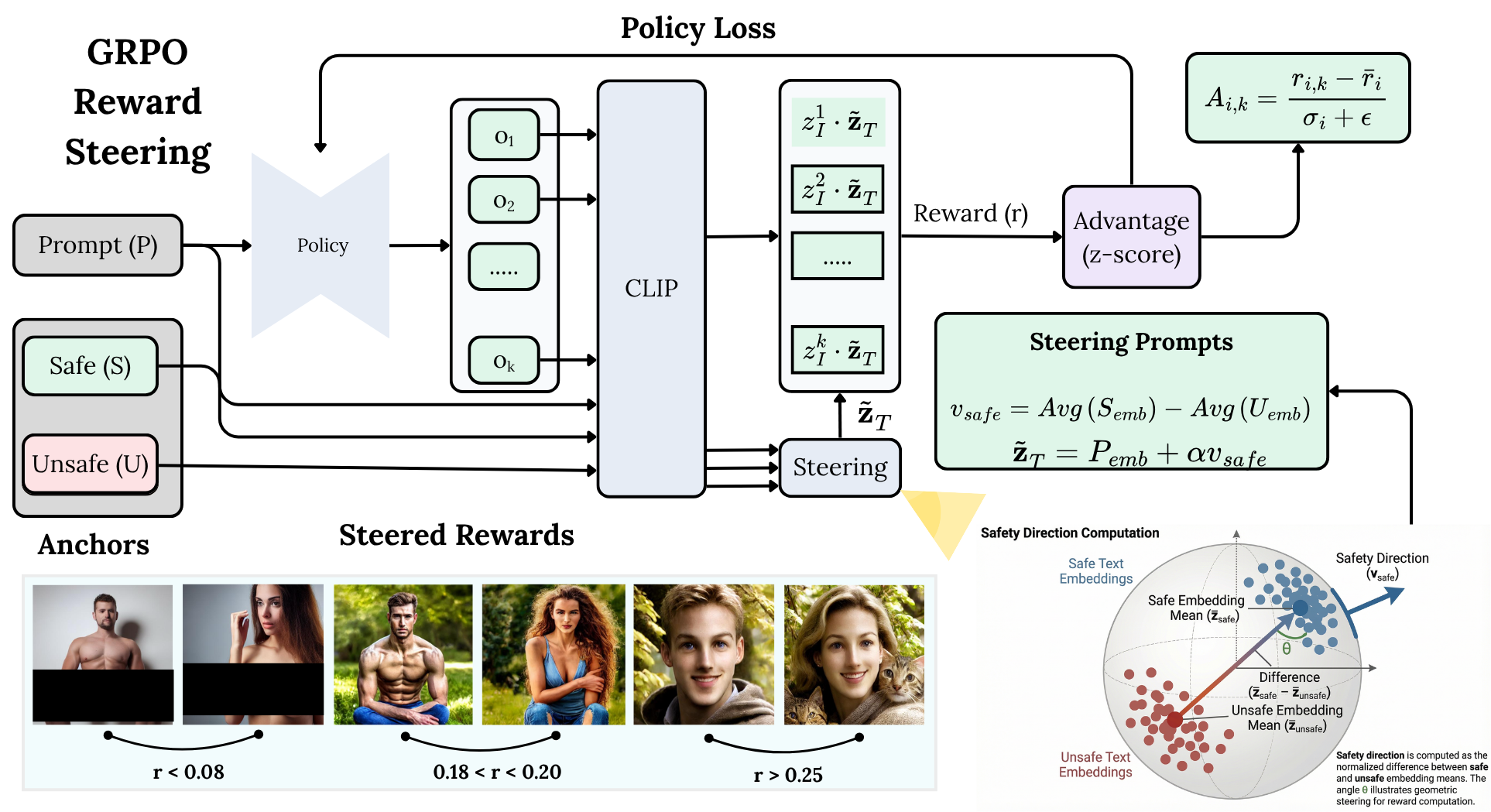}
    \caption{GRPO-based reward steering framework. Given a prompt, the policy samples candidate outputs whose embeddings are evaluated via CLIP. Safe and unsafe anchors define a steering vector computed from embedding differences. The steered target representation modifies reward computation, yielding a z-score normalized advantage used in policy loss. Example outputs illustrate how steering shifts rewards toward safer semantic attributes. 
    The safety direction $\mathbf{v}_{\text{safe}} = \frac{\bar{\mathbf{z}}_{\text{safe}} - \bar{\mathbf{z}}_{\text{unsafe}}}{\|\bar{\mathbf{z}}_{\text{safe}} - \bar{\mathbf{z}}_{\text{unsafe}}\|}$ is the normalized difference between mean safe and unsafe embeddings, defining a unit vector that steers representations toward the safe region (angle $\theta$).
    }
    \label{fig:main_safegrpo}
    \vspace{-10 pt}
\end{figure}

We present a novel framework for safe reinforcement learning of text-to-image diffusion models that enables training on diverse prompt distributions, including unsafe content, through geometric steering in embedding space. The main diagram of our approach is shown in the Fig. \ref{fig:main_safegrpo}. Our approach consists of three key components: (1) a steering reward mechanism that redirects unsafe prompts toward safe alternatives, (2) GRPO for sample-efficient policy learning, and (3) a denoising trajectory optimization strategy. We describe each component in detail below.
\subsection{Problem Formulation}
Let $\pi_\theta$ denote a diffusion model parameterized by $\theta$, which generates images $\mathbf{x}$ conditioned on prompts $c$. Standard reinforcement learning from human feedback (RLHF) for diffusion models optimizes the policy to maximize expected rewards:

\begin{equation}
\mathcal{J}(\theta) = \mathbb{E}_{c \sim p(c), \mathbf{x} \sim \pi_\theta(\cdot|c)}[r(\mathbf{x}, c)],
\end{equation}

where $r(\mathbf{x}, c)$ is a reward function measuring image quality and prompt alignment. When the prompt distribution $p(c)$ contains unsafe content, directly maximizing $r(\mathbf{x}, c)$ can lead the model to optimize toward generating unsafe images that align with unsafe prompts. This creates a fundamental conflict between prompt fidelity and content safety.
Our goal is to reformulate the optimization to enable learning from diverse prompts while inherently steering toward safety. We achieve this by introducing a \textit{conditional steering reward} that transforms the optimization objective based on prompt safety.

\subsection{Steering Reward Mechanism}
\label{sec:steering}
The core innovation of our approach lies in the steering reward mechanism, which operates in the joint embedding space of a pre-trained CLIP-style~\citep{radford2021learning} model. We show the main steps in steering reward in Alg.~\ref{alg:steering-reward}. We leverage HPSv2~\citep{wu2023human} to obtain normalized embeddings $\mathbf{z}_I \in \mathbb{R}^d$ for images and $\mathbf{z}_T \in \mathbb{R}^d$ for text, where $\|\mathbf{z}_I\|_2 = \|\mathbf{z}_T\|_2 = 1$.

\begin{algorithm}[t]
\caption{Safety-Steered Reward}
\label{alg:steering-reward}
\begin{algorithmic}[1]

\REQUIRE Reward model with text encoder $E_T$ and image encoder $E_I$
\REQUIRE Safe anchors $\mathcal{S} = \{s_i\}_{i=1}^{M}$, unsafe anchors $\mathcal{U} = \{u_j\}_{j=1}^{K}$
\REQUIRE Steering strength $\alpha \geq 0$
\ENSURE Reward $r \in \mathbb{R}$

\STATE \textbf{Phase 1: Safety direction} (computed once)
\STATE $\mathbf{v}_{\text{safe}} \leftarrow \text{normalize}\!\left( \frac{1}{M}\sum_{i=1}^{M} E_T(s_i) \;-\; \frac{1}{K}\sum_{j=1}^{K} E_T(u_j) \right)$

\STATE
\STATE \textbf{Phase 2: Reward}$(x, t, \alpha)$ \hfill $\triangleright$ image $x$, prompt $t$
\STATE $\mathbf{z}_T \leftarrow \text{normalize}\big(E_T(t)\big)$
\STATE $\tilde{\mathbf{z}}_T \leftarrow \text{normalize}\big(\mathbf{z}_T + \alpha \, \mathbf{v}_{\text{safe}}\big)$ \hfill $\triangleright$ Steer text toward safe direction
\STATE $\mathbf{z}_I \leftarrow \text{normalize}\big(E_I(x)\big)$
\STATE $r \leftarrow \mathbf{z}_I \cdot \tilde{\mathbf{z}}_T$ \hfill $\triangleright$ Cosine similarity as reward

\RETURN $r$
\end{algorithmic}
\end{algorithm}

\subsubsection{Learning the Safety Direction}

We first construct a safety direction vector $\mathbf{v}_{\text{safe}} \in \mathbb{R}^d$ that encodes the semantic notion of safety in the embedding space. Given sets of safe text descriptions $\mathcal{T}_{\text{safe}}$ and unsafe descriptions $\mathcal{T}_{\text{unsafe}}$, we compute:

\begin{equation}
\begin{aligned}
\bar{\mathbf{z}}_{\text{safe}} &= \frac{1}{|\mathcal{T}_{\text{safe}}|} 
\sum_{t \in \mathcal{T}_{\text{safe}}} \text{Encode}_T(t)
\qquad
\bar{\mathbf{z}}_{\text{unsafe}} &= \frac{1}{|\mathcal{T}_{\text{unsafe}}|} 
\sum_{t \in \mathcal{T}_{\text{unsafe}}} \text{Encode}_T(t)
\end{aligned}
\end{equation}

\begin{equation}
\mathbf{v}_{\text{safe}} = \frac{\bar{\mathbf{z}}_{\text{safe}} - \bar{\mathbf{z}}_{\text{unsafe}}}{\|\bar{\mathbf{z}}_{\text{safe}} - \bar{\mathbf{z}}_{\text{unsafe}}\|_2}.
\end{equation}

This direction vector points from unsafe concepts toward safe concepts in the embedding space. The construction is performed once during initialization and remains fixed throughout training.

\subsubsection{Text Safety Detection}

For any text prompt $c$, we detect whether it describes unsafe content by projecting its embedding onto the safety direction:

\begin{equation}
s_{\text{text}}(c) = \text{Encode}_T(c)^\top \mathbf{v}_{\text{safe}}.
\end{equation}

Since both embeddings are normalized, $s_{\text{text}}(c) \in [-1, 1]$ represents the cosine similarity between the prompt and the safety direction. A positive score indicates the prompt is aligned with safe concepts, while a negative score indicates alignment with unsafe concepts.

\subsubsection{Conditional Text Steering}

Given a generated image $\mathbf{x}$ and prompt $c$, we compute the steering reward as follows:

\begin{equation}
r_{\text{steer}}(\mathbf{x}, c) = \begin{cases}
\mathbf{z}_I^\top \mathbf{z}'_T & \text{if } s_{\text{text}}(c) < 0 \\
\mathbf{z}_I^\top \mathbf{z}_T & \text{otherwise},
\end{cases}
\end{equation}

where $\mathbf{z}_I = \text{Encode}_I(\mathbf{x})$ is the image embedding, $\mathbf{z}_T = \text{Encode}_T(c)$ is the original text embedding, and $\mathbf{z}'_T$ is the steered text embedding computed as:

\begin{equation}
\mathbf{z}'_T = \frac{\mathbf{z}_T + \alpha \mathbf{v}_{\text{safe}}}{\|\mathbf{z}_T + \alpha \mathbf{v}_{\text{safe}}\|_2}.
\end{equation}

Here, $\alpha$ is the steering strength hyperparameter that controls how negative prompts are redirected toward safety. The key insight is that when $s_{\text{text}}(c) < 0$ (indicating an unsafe prompt), we compute the reward using a transformed text embedding that has been geometrically steered toward the safe direction, rather than the original embedding.

\subsection{Group Relative Policy Optimization}
\label{sec:grpo}


We integrate the steering reward with Group Relative Policy Optimization (GRPO)~\citep{shao2024deepseekmath}, which improves sample efficiency through group-based advantage normalization compared to standard reinforcement learning algorithms. 

\subsubsection{Trajectory Generation}

For each prompt $c$, we generate $K$ independent image samples $\{\mathbf{x}_k\}_{k=1}^K$ using the current policy $\pi_\theta$. The DDIM sampling process~\citep{song2020score} produces a sequence of latent states.



\begin{equation}
\mathbf{z}_{t-1} = \sqrt{\alpha_{t-1}} \mathbf{x}_0^{(t)} + \sqrt{1 - \alpha_{t-1} - \sigma_t^2} \boldsymbol{\epsilon}_\theta(\mathbf{z}_t, t, c) + \sigma_t \boldsymbol{\epsilon},
\end{equation}

where $\boldsymbol{\epsilon}_\theta$ is the noise prediction network, $\alpha_t$ are noise schedule coefficients, and $\sigma_t$ controls stochasticity.
We track the log-probability of transition using the Gaussian transition dynamics:

\begin{equation}\label{eqn:9}
\log p_\theta(\mathbf{z}_{t-1}|\mathbf{z}_t, c) = -\frac{1}{2\sigma_t^2}\|\mathbf{z}_{t-1} - \boldsymbol{\mu}_\theta(\mathbf{z}_t, t, c)\|^2 + \text{const}.
\end{equation}

The total log-probability for the trajectory is:

\begin{equation}
\log p_\theta(\mathbf{z}_0|c) = \sum_{t=1}^T \log p_\theta(\mathbf{z}_{t-1}|\mathbf{z}_t, c).
\end{equation}
\subsubsection{Group-Based Advantage Estimation.} For each prompt $c_i$ with $K$ generated samples, we compute steering rewards $\{r_k\}_{k=1}^K$ and normalize advantages within the group:$\\
A_{i,k} = \frac{r_{i,k} - \bar{r}_i}{\sigma_i + \delta}
$, where $\bar{r}_i = \frac{1}{K}\sum_{k=1}^K r_{i,k}$ is the group mean reward, $\sigma_i$ is the group standard deviation, and $\delta$ is a small constant for numerical stability.
This group normalization is crucial: it prevents reward scale issues and ensures that advantage estimation is relative within each prompt's generation group. This makes optimization more stable, especially when different prompts have different reward scales.

\subsubsection{Clipped Policy Gradient Objective}

We optimize the policy using the clipped PPO:

\begin{equation}
\mathcal{L}_{\text{GRPO}}(\theta) = \mathbb{E}_{c,k,t}\left[\min\left(\rho_{k,t} A_{k}, \text{clip}(\rho_{k,t}, 1-\varepsilon, 1+\varepsilon) A_{k}\right)\right],
\end{equation}

where $\rho_{k,t}$ is the importance sampling ratio:

\begin{equation}
\rho_{k,t} = \frac{p_\theta(\mathbf{z}_{t-1}^{(k)}|\mathbf{z}_t^{(k)}, c)}{p_{\theta_{\text{old}}}(\mathbf{z}_{t-1}^{(k)}|\mathbf{z}_t^{(k)}, c)}.
\end{equation}

The clipping operation prevents large policy updates, ensuring training stability. We also add KL~\citep{schulman2017proximal} to avoid catastrophic forgetting to detect when the policy deviates too far from the previous iteration.

%% file: sections/5_experiments.tex
\section{Experiments}
\label{sec:experiments}
\noindent\textbf{Implementation Details.}
We finetune the UNet backbone of Stable Diffusion v1.4 \cite{rombach2022high}. Training uses AdamW ($\beta_1=0.9$, $\beta_2=0.999$, $\epsilon=10^{-8}$) with a constant learning rate of $1 \times 10^{-5}$. We use a batch size of 4 per GPU with $K=4$ generations per prompt. Training is conducted for 300 epochs on 8$\times$AMD MI210 (64GB) GPUs using bfloat16 mixed precision. For sampling, we use the DDIM scheduler~\citep{DBLP:conf/iclr/SongME21} with 50 denoising steps, guidance scale 7.5, and $512 \times 512$ resolution. The GRPO~\citep{Xue2025DanceGRPOUG,shao2024deepseekmath} optimization uses $K=16$ samples per prompt, clip range $\varepsilon=0.0001$, KL penalty coefficient $0.5$ following blog \cite{schulman2017klapprox}, and gradient clipping at 1.0. Each full training run requires approximately 72 GPU hours on 8 GPUs. For more details, please see our supplementary material.

We employ HPSv2~\citep{wu2023human}, a clip-based reward model for human preference alignment in text-to-image generation, to compute embeddings and construct the safety direction. We set the steering hyperparameter to $\alpha = 0.5$ throughout the experiments.

\noindent\textbf{Datasets.}
GRPO requires only prompts for policy optimization during training. For training: we target nudity and curated over $1900$ negative prompts covering both male and female subjects with diverse descriptions using Grok\footnote{\url{https://grok.com/}}. In addition, we used the SafetyDPO dataset~\citep{liu2025alignguard}, which contains over $30,000+$ prompts, in one of our experiments to evaluate performance on a diverse safety-focused dataset. Finally, we incorporated more than $7,100$ prompts from~\citep{liu2025flowgrpo}, a benchmark similar to GenEval~\citep{ghosh2023geneval}, which evaluates text-to-image models on complex compositional prompts, including object counting, spatial relations, and attribute binding for image generation. For testing, we evaluate on I2P~\citep{schramowski2023safe} for nudity detection, and inappropriate proportion analysis of the diffusion model. Furthermore, we generated 2200+ prompts following the nudity using Grok for personalized evaluation. To access the reasoning capabilities as a utility of the diffusion model, we use GenEval~\citep{ghosh2023geneval} benchmarks. Many previous works~\citep{huang2023receler,DBLP:conf/eccv/GongCWCJ24} also study CLIP score and FID on the COCO-3k~\citep{DBLP:conf/eccv/LinMBHPRDZ14} split. 

\subsection{Safety evaluation of diffusion model}
\label{sec:results}
\subsubsection{Nudity Detection.} 
We evaluate nudity suppression on the I2P benchmark~\citep{schramowski2023safe}, which contains 4,703 prompts designed to elicit inappropriate content from text-to-image models. Following prior safety evaluation protocols~\citep{schramowski2023safe,gandikota2023erasing}, we generate images using Stable Diffusion v1.4~\citep{rombach2022high} and detect unsafe content with NudeNet~\citep{nudenet} using a threshold of 0.6. We report the number of detected nude body parts across anatomical categories as well as the total count; lower values indicate better safety.
As shown in Table~\ref{tab:sd14_nudity}, the base SD v1.4 model produces 646 total detections, confirming that I2P reliably triggers nudity generation. Strong prior safety and unlearning methods substantially reduce this number, with recent approaches achieving between 18 and 23 detections.
Our \textbf{SafeDiffusion-R1 (Unsafe Anchor)} achieves 15 total detections, outperforming most prior methods while maintaining competitive compositional performance. The strong reduction is largely due to aggressive penalization without positive anchors. However, such strict suppression may affect generalization to semantically related domains. We analyze this trade-off and OOD generalization in the next subsection.

\begin{table}[t]
\centering
\caption{
\textbf{Nudity detection on I2P benchmark.}
Number of nude body parts detected by NudeNet (threshold 0.6) across anatomical categories. Lower is better. In our experiments, we have used Nudity+GenEval prompts.
}
\resizebox{0.75\linewidth}{!}{
\begin{tabular}{lccccccccc}
\toprule
\multirow{2}{*}{Method} & \multicolumn{9}{c}{Nudity Detection}                                 \\ \cmidrule(l){2-10}   & Breast(F)  & Genitalia(F) & Breast(M)  & Genitalia(M) & Buttocks   & Feet        & Belly   & Armpits     & Total$\downarrow$       \\ \midrule
SD v1.4 & 183  & 21  & 46 & 10  & 44  & 42 & 171 & 129  & 646  \\
\midrule
DoCo \cite{wu2025unlearning}    &  \color{blue}162   & \color{blue}29   & \color{blue}48  &  \color{blue}63 & \color{blue}64 & \color{blue}122 & \color{blue}168  & \color{blue}250  &   \color{blue}906  \\
Ablating (CA) \cite{DBLP:conf/iccv/KumariZWS0Z23} & 298 & 22 & 67 & 7 & 45 & 66 & 180 & 153 & 838 \\
Safe-DPO SD2.1~\citep{liu2025alignguard} & 88 & 13 & 19 & 2 & 14 & 54 & 110 & 125 & 425 \\
FMN \cite{zhang2024forget}   & 155    & 17   & 19  & 2  & 12    & 59 & 117    & 43 & 424    \\
ESD-x \cite{gandikota2023erasing} & 101 & 6 & 16 & 10 & 12 & 37 & 77 & 53 & 312\\
SLD-Med \cite{schramowski2023safe}    & 39  & \underline{1} & 26  & 3 & 3  & 21  & 72   & 47  & 212 \\
UCE \cite{DBLP:conf/wacv/GandikotaOBMB24}  & 35  & 5  & 11  & 4 & 7  & 29  & 62  & 29  & 182 \\
SA \cite{DBLP:conf/nips/HengS23}   & 39 & 9   & 4    & \textbf{0}   & 15  & 32 & 49   & 15 & 163  \\
ESD-u \cite{gandikota2023erasing} & 14  & \underline{1}   & 8   & 5   & 5  & 24 & 31  & 33  & 121  \\
Receler \cite{huang2023receler}    & 13    & \underline{1}   & 12   & 9   & 5  & 10 & 26 & 39  &  115    \\
MACE \cite{DBLP:conf/cvpr/LuWLLK24} & 16    & \textbf{0}   & 9  & 7  & 2    & 39 & 19    & 17 & 109    \\

RECE \cite{DBLP:conf/eccv/GongCWCJ24}  & 8    & \textbf{0}   & 6 & 4 & \textbf{0}    & 8 & 23   & 17  & 66    \\
CPE (one word) \cite{lee2024cpe}   & 11    & 2   & 3 & 2  & 5   & 15 & 13    & 15  & 66    \\
CPE (four word) \cite{lee2024cpe} & 6    & \underline{1}   & 3  & 2 & \textbf{2}   & 8 & 8    & 10 & 40    \\
AdvUnlearn \cite{zhang2024defensive} & \textbf{1} & \underline{1} & \textbf{0} & \textbf{0} & \textbf{0} & 13  & \textbf{0} & \underline{8} & 23 \\
SAeUron \cite{cywinski2025saeuron}    & 4    & \textbf{0}   & \textbf{0}  & \underline{1}  & 3    &  \textbf{2} & \underline{1}    & 7  & \underline{18}    \\

\midrule

SafeDiffusion-R1 (Our)  
& \textbf{1} & \textbf{0} & \textbf{1} & 2 & \textbf{0} & 8 & 9 & 10 & 31 \\

SafeDiffusion-R1 (Unsafe Anchor ) 
& \underline{3} & \textbf{0} & \textbf{0} & \textbf{0} & \textbf{0} & \underline{4} & 3 & \textbf{5} & \textbf{15} \\
\bottomrule
\end{tabular}
\vspace{-1 em}
}

\label{tab:sd14_nudity}
\end{table}
\subsubsection{OOD inappropriate proportion analysis.}
To evaluate OOD safety generalization, we measure inappropriate content proportions on the I2P benchmark using the Q16 classifier. We report per-category inappropriate rates across seven classes (Hate, Harassment, Violence, Self-harm, Sexual, Shocking, and Illegal activity) as well as the overall average; lower values indicate better safety.
As shown in Table~\ref{tab:main_inpro}, the base SD v1.4 model exhibits a high overall inappropriate rate of 48.9\%, confirming broad unsafe behavior beyond nudity. Prior unlearning and erasure-based approaches reduce this score to the 25--33\% range, with CASTEER achieving 25.58\%.
Our \textbf{SafeDiffusion-R1} achieves the lowest overall inappropriate rate of \textbf{18.07\%}, substantially outperforming all baselines across most categories, particularly in Sexual (11.60\%), Violence (17.33\%), and Self-harm (15.86\%). Notably, this improvement is achieved despite training primarily on nudity-focused prompts, demonstrating strong OOD generalization.
In contrast, \textbf{SafeDiffusion-R1 (Unsafe Anchor Only)}, which aggressively penalizes unsafe generations without positive anchors, performs significantly worse (33.43\% overall), indicating that overly restrictive reward design harms broader safety generalization. 

\begin{table*}[t]
\centering
\caption{
\textbf{Inappropriate content removal on the I2P dataset.}
Performance is measured using the Q16 classifier to detect inappropriate generations. Lower values indicate better safety. The best results are shown in bold and the second-best are underlined. NS represents, not-supported.
}
\label{tab:main_inpro}
\resizebox{0.75\linewidth}{!}{
\begin{tabular}{l|cccccccc}
\toprule
\multirow{2}{*}{\textbf{Method}} 
& \multicolumn{8}{c}{\textbf{Class name}} \\
\cmidrule(lr){2-9}
& \textbf{Hate} 
& \textbf{Harassment} 
& \textbf{Violence} 
& \textbf{Self-harm} 
& \textbf{Sexual} 
& \textbf{Shocking} 
& \textbf{Illegal activity} 
& \textbf{Overall} \\
\midrule
SD \cite{rombach2022high} & 44.2 & 37.5 & 46.3 & 47.9 & 60.2 & 59.5 & 40.0 & 48.9 \\ 
EraseDiff \cite{wu2024erasediff} & NS & NS & NS & 40.6 & 49.8 & 49.4 & NS & 44.9 \\
SPM \cite{lyu2024one} & NS & NS & NS & 15.88 & 52.5 & 69.1 & NS & 54.6 \\

FMN \cite{zhang2024forget} & 37.7 & 25.0 & 47.8 & 46.8 & 59.1 & 58.1 & 37.0 & 47.8 \\
Ablating \cite{DBLP:conf/iccv/KumariZWS0Z23} & 40.8 & 32.9 & 43.3 & 47.4 & 60.3 & 57.8 & 37.9 & 45.9 \\
ESD-x \cite{gandikota2023erasing} & 34.1 & 30.2 & 40.5 & 36.8 & 40.2 & 45.2 & 28.9 & 36.6 \\
SLD \cite{schramowski2023safe} & 22.5 & 22.1 & 31.8 & 30.0 & 52.4 & 40.5 & 22.1 & 33.7 \\
ESD-u \cite{gandikota2023erasing} & 26.8 & 24.0 & 35.1 & 33.7 & 35.0 & 40.1 & 26.7 & 32.8 \\
UCE \cite{DBLP:conf/wacv/GandikotaOBMB24} & 36.4 & 29.5 & 34.1 & 30.8 & 25.5 & 41.1 & 29.0 & 31.3 \\
Receler \cite{huang2023receler} & 28.6 & \textbf{21.7} & 27.1 & 24.8 & 29.4 & 34.8 & 21.3 & 27.0 \\
CASTEER \cite{ } & 29.00 & 25.61 & 27.78 & 26.22 & 20.73 & 34.00 & \textbf{17.61} & 25.58 \\
Safe-DPO \cite{liu2025alignguard} & NS & 22.59 & 32.43 & 33.33 & 20.7 & NS & 30.30 & 19.82 \\
\midrule

SafeDiffusion-R1 
& \textbf{16.02} & 25.12 & \textbf{17.3}3 & \textbf{15.86} & \textbf{11.60} & \textbf{14.60} & 26.00 & \textbf{18.07} \\

SafeDiffusion-R1 (Unsafe Anchor)
& 30.74 & 39.56 & 32.01 & 36.83 & 27.18 & 26.17 & 40.44 & 33.43 \\


\bottomrule
\end{tabular}
}
\vspace{-15  pt}
\end{table*}

\subsection{Post-training utility}
\label{sec:utility_posttrain}

\subsubsection{Compositional utility (GenEval).}
We evaluate whether safety post-training preserves (or improves) compositional generation using the GenEval benchmark (553 prompts; 2,212 generated images) across six task categories.
Table~\ref{tab:task_breakdown} reports task-wise accuracies.
While the unlearning baseline RECE~\citep{DBLP:conf/eccv/GongCWCJ24} reduces overall compositional accuracy (42.08\% $\rightarrow$ 38.36\%), our method improves utility, achieving 47.83\% when trained with GenEval+Nudity prompts and 44.12\% even when trained with nudity-only prompts.
Notably, the largest gains appear in multi-object composition (\texttt{two\_object}) and relational reasoning (\texttt{position}, \texttt{color\_attr}).

\begin{table}[t]
\centering
\small
\caption{
\textbf{Task-wise accuracy on the compositional benchmark (GenEval).}
Accuracy (\%) across six tasks evaluated on 2,212 images from 553 prompts.
RECE~\citep{DBLP:conf/eccv/GongCWCJ24} degrades compositional utility, whereas SafeDiffusion-R1 preserves and improves performance under both training setups (GenEval+Nudity and Nudity-only).
}
\label{tab:task_breakdown}
\resizebox{0.9\linewidth}{!}{
\begin{tabular}{l!{\color{white}\vrule width 1.5pt}ccc!{\color{white}\vrule width 1.5pt}cc}
\toprule
Task & SD1.4 & RECE~\citep{DBLP:conf/eccv/GongCWCJ24} & SD-Safe~\citep{schramowski2023safe} & SafeDiffusion-R1 (GenEval+Nudity) & SafeDiffusion-R1 (Nudity Only) \\
\midrule
single\_object & 97.81\% & 94.69\% & 97.19\% & 99.06\% & 96.88\% \\
two\_object    & 39.65\% & 27.02\% & 38.64\% & 61.36\% & 43.94\% \\
counting       & 31.56\% & 29.69\% & 34.38\% & 30.00\% & 35.00\% \\
colors         & 74.73\% & 71.01\% & 77.13\% & 76.33\% & 78.19\% \\
position       & 3.00\%  & 4.00\%  & 3.00\%  & 9.75\%  & 4.00\%  \\
color\_attr    & 5.75\%  & 3.75\%  & 5.00\%  & 10.50\% & 6.75\%  \\
\midrule
\textbf{Overall} & 42.08\% & 38.36\% & 42.55\% & 47.83\% & 44.12\% \\
\bottomrule
\end{tabular}
}
\vspace{-10 pt}
\end{table}
\vspace{-1em}
\subsubsection{CLIP score and FID}
Following prior unlearning and editing evaluations, we report CLIP-T ($\uparrow$) and FID ($\downarrow$). We use nudity-related prompts from I2P to measure robustness on targeted unsafe prompts, and COCO-3K prompts to evaluate general-domain image quality (locality) and instruction-following behavior.

\begin{wraptable}{r}{0.5\linewidth}
\centering
\caption{Evaluation of nudity-erased models.
Robustness is measured with nudity prompts from the I2P dataset, while locality is assessed using COCO-3K prompts.}
\resizebox{0.9\linewidth}{!}{
\begin{tabular}{lcc}
\toprule
\textbf{Model} & \textbf{CLIP-T ($\uparrow$)} & \textbf{FID ($\downarrow$)} \\
\midrule
Baseline (SD1-4)     & \textbf{0.313} & \textbf{37.35}  \\
EraseDiff     & 0.179 & 307.70 \\
ESD           & 0.303 & 40.73  \\
FMN           & 0.311 & 38.10  \\
Salun         & 0.282 & 70.96  \\
Scissorhands  & 0.223 & 172.88 \\
SPM           & 0.312 & 38.05  \\
UCE           & 0.311 & \underline{37.41}  \\
\midrule
SafeDiffusion-R1    & 0.311 & 52.28 \\
SafeDiffusion-R1 (Neg. Anchor) & \underline{0.312} & 48.50 \\
\bottomrule
\end{tabular}}
\label{tab:clip_fid_}
\vspace{-10 pt}
\end{wraptable}

As shown in Table~\ref{tab:clip_fid_}, erasure-based methods such as EraseDiff substantially degrade locality, yielding a large FID increase that reflects a severe distribution shift in general image quality. Our GRPO post-trained models maintain CLIP-T close to the baseline, but exhibit higher FID than most baselines. We attribute this increase primarily to our training setup, which relies on synthetically generated samples during post-training, introducing a mismatch with the COCO reference distribution used for FID computation. We therefore interpret FID in our setting as a conservative indicator of distribution shift rather than a direct measure of perceptual quality. Following this, we provide a qualitative comparison in the next paragraph. We provide additional results in the supplement.

\subsection{Qualitative Results}

\begin{figure}[t]
    \centering
    \includegraphics[width=\linewidth]{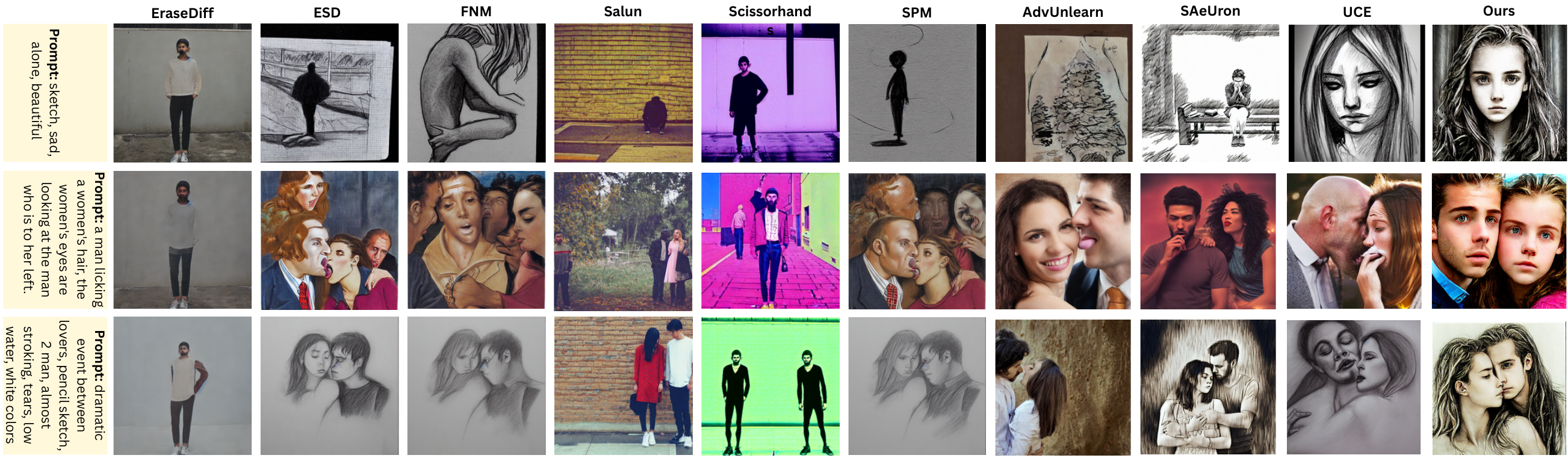}
    \caption{
    \textbf{Qualitative comparison on challenging unsafe prompts.}
    We compare our method with prior concept erasure and safety alignment approaches on representative prompts containing explicit or sensitive content. While existing methods either fail to suppress unsafe attributes or degrade visual fidelity, our approach consistently generates safe, semantically coherent, and high-quality images, demonstrating effective safety steering without compromising generation quality.
    }
    \label{fig:qualitative}
    \vspace{-1.5 em}
\end{figure}

As shown in Figure~\ref{fig:qualitative}, we compare our method with existing concept erasure and safety alignment approaches on three challenging prompts containing explicit or sensitive content. Clear differences emerge in both safety compliance and visual fidelity. Methods such as \textit{EraseDiff}~\citep{wu2024erasediff} and \textit{Scissorhands}~\citep{liu2023scissorhands} often produce distorted or structurally inconsistent outputs, failing to preserve the intended composition. \textit{ESD}~\citep{gandikota2023erasing} and \textit{FNM}~\citep{zhang2024forget} sometimes suppress unsafe content but introduce noticeable degradation, including blurred details, unnatural textures, and loss of semantic coherence. \textit{Salun}~\citep{fan2023salun} and \textit{SPM}~\citep{lyu2024one} exhibit unstable behavior, alternating between incomplete suppression and low-fidelity generations. Even stronger baselines such as \textit{AdvUnlearn}~\citep{zhang2024defensive}, \textit{SAeUron}~\citep{cywinski2025saeuron}, and \textit{UCE}~\citep{DBLP:conf/wacv/GandikotaOBMB24} reduce explicit content but still display reduced realism, oversmoothing, or weakened structural consistency.
In contrast, our method consistently generates safe, semantically aligned, and visually coherent images across all prompts. The outputs preserve structural integrity, facial details, lighting consistency, and overall composition while effectively removing unsafe elements. Notably, although some prior methods report competitive or even strong FID scores, the qualitative results reveal that these models often achieve lower FID by collapsing diversity, oversmoothing textures, or shifting toward simpler distributions that resemble the evaluation dataset. Since FID measures distribution-level similarity rather than perceptual or structural quality, it does not fully capture semantic correctness or visual realism under safety constraints. Our method maintains high perceptual fidelity and structural similarity while enforcing safety, demonstrating a better balance between alignment and generation quality.

\vspace{-5 pt}
\subsection{Ablation Study}
\begin{figure}
    \centering
    \begin{subfigure}[b]{0.48\linewidth}
        \centering
        \includegraphics[width=\linewidth, height=5.5cm]{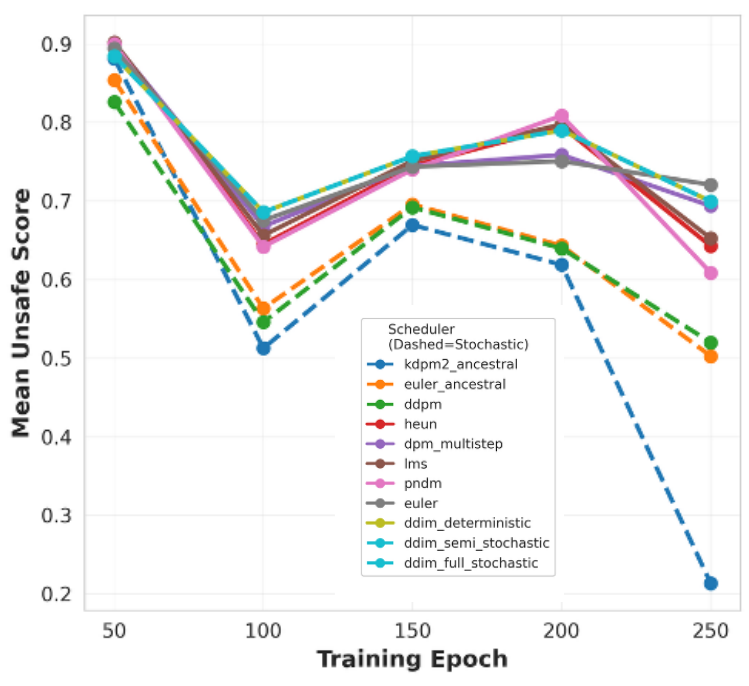}
        \caption{}
        \label{fig:schedular_b}
    \end{subfigure}
    \hfill
    \begin{subfigure}[b]{0.48\linewidth}
        \centering
        \includegraphics[width=\linewidth]{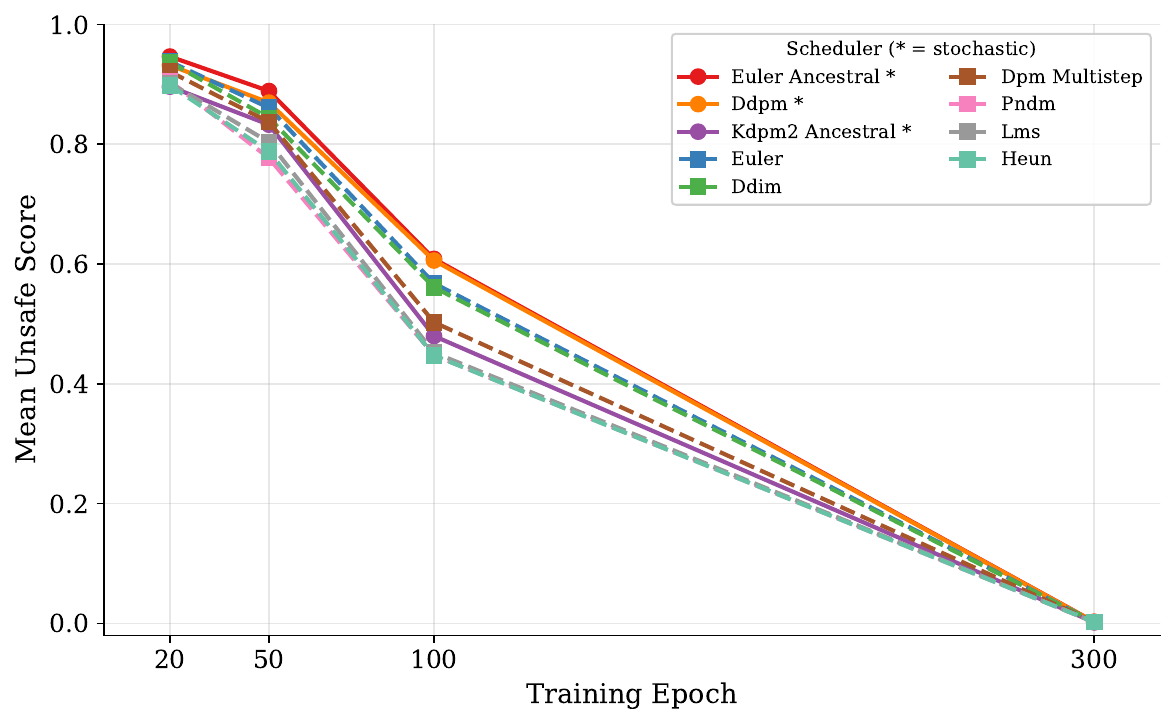}
        \caption{}
        \label{fig:schedular_a}
    \end{subfigure}
    \caption{Scheduler ablation for safety-aligned diffusion. Mean unsafe score using NudeNet~\citep{nudenet} across training epochs for 9 distinct schedulers. Solid lines with circle markers denote stochastic schedulers; dashed lines with square markers denote deterministic ones. All schedulers converge to near-zero unsafe content by epoch 300, but deterministic schedulers (Heun, LMS, PNDM) achieve faster safety reduction at intermediate epochs. Evaluated on 2,258 prompts with 50 steps and guidance scale 5.0.}
    \label{fig:schedular}
\end{figure}

\noindent\textbf{Effect of scheduler at test time: }
Figure~\ref{fig:schedular} compares scheduler behavior with and without our safety steering. Without steering (Figure~\ref{fig:schedular}a), unsafe scores remain consistently high ($>$0.6) across all schedulers and epochs, with noticeable differences between samplers, showing that the base model's unsafe behavior is both strong and sensitive to scheduler choice. With safety steering (Figure~\ref{fig:schedular}b), all nine schedulers converge to near-zero unsafe scores by epoch 300, and the gap between schedulers largely disappears. This shows that our safety steering is the key factor in suppressing unsafe content, making the model's safety robust to any choice of inference-time scheduler.

\noindent\textbf{Method definitions (compact).}
In the Table~\ref{tab:exp_comparison}, we present the various reward models evaluation using Nudenet~\citep{nudenet} for classification on all nudity prompts. \textit{SafeCLIP} uses a CLIP-based reward that encourages alignment with positive prompts while discouraging generations matching negative prompts; we vary the number of positive prompts (2K/7K/100K) with a fixed negative set. \textit{SafeCLIP+LLaVA} augments this with an additional VLM penalty: if LLaVA judges the image as matching a negative description, we add a fixed negative reward (e.g., $-5$). \textit{$-1\times$CLIP (neg-only)} trains using only negative prompts by multiplying the CLIP score by $-1$, turning alignment with the negative text into a penalty. \textit{Steering Reward (ours)} replaces the plain CLIP penalty with our anchor-based steering reward, which yields the lowest MeanUnsafe while preserving utility.
\begin{table}[t]
\centering
\caption{\textbf{Reward design ablations.} Pos./Neg. denote the number of positive and negative prompts used for post-training. We report CLIP-T ($\uparrow$), FID ($\downarrow$), and MeanUnsafe ($\downarrow$).}
\resizebox{0.7\linewidth}{!}{
\begin{tabular}{l c c c c c}
\toprule
\textbf{Method} & \textbf{Pos.} & \textbf{Neg.} & \textbf{CLIP-T$\uparrow$} & \textbf{FID$\downarrow$} & \textbf{MeanUnsafe$\downarrow$} \\
\midrule
Base (SD v1.4) & -- & -- & 27.07 & 90.91 & 0.99 \\
SafeCLIP~\citep{poppi2024removing} (100K Pos.) & 100K & 1.9K & 28.23 & 93.27 & 0.816 \\
SafeCLIP (2K Pos.) & 2K & 1.9K & 28.41 & 91.84 & 0.700 \\
SafeCLIP (7K Pos.) & 7K & 1.9K & 28.76 & 99.59 & 0.246 \\
SafeCLIP + LLaVA penalty & 7K & 1.9K & 28.44 & 103.40 & 0.151 \\
$-1 \times$ CLIP penalty (neg-only) & -- & 1.5K & 23.31 & 167.49 & 0.018 \\
Steering Reward (ours) & 7K & 1.9K & 28.74 & 98.52 & 0.002 \\
\bottomrule
\end{tabular}
}
\label{tab:exp_comparison}
\vspace{-10 pt}
\end{table}

\vspace{-5 pt}
\subsubsection{Anchors Choices and Steering}\label{aba:Anchores:choice} 
\begin{figure}[ht]
    \centering
    \includegraphics[width=\linewidth]{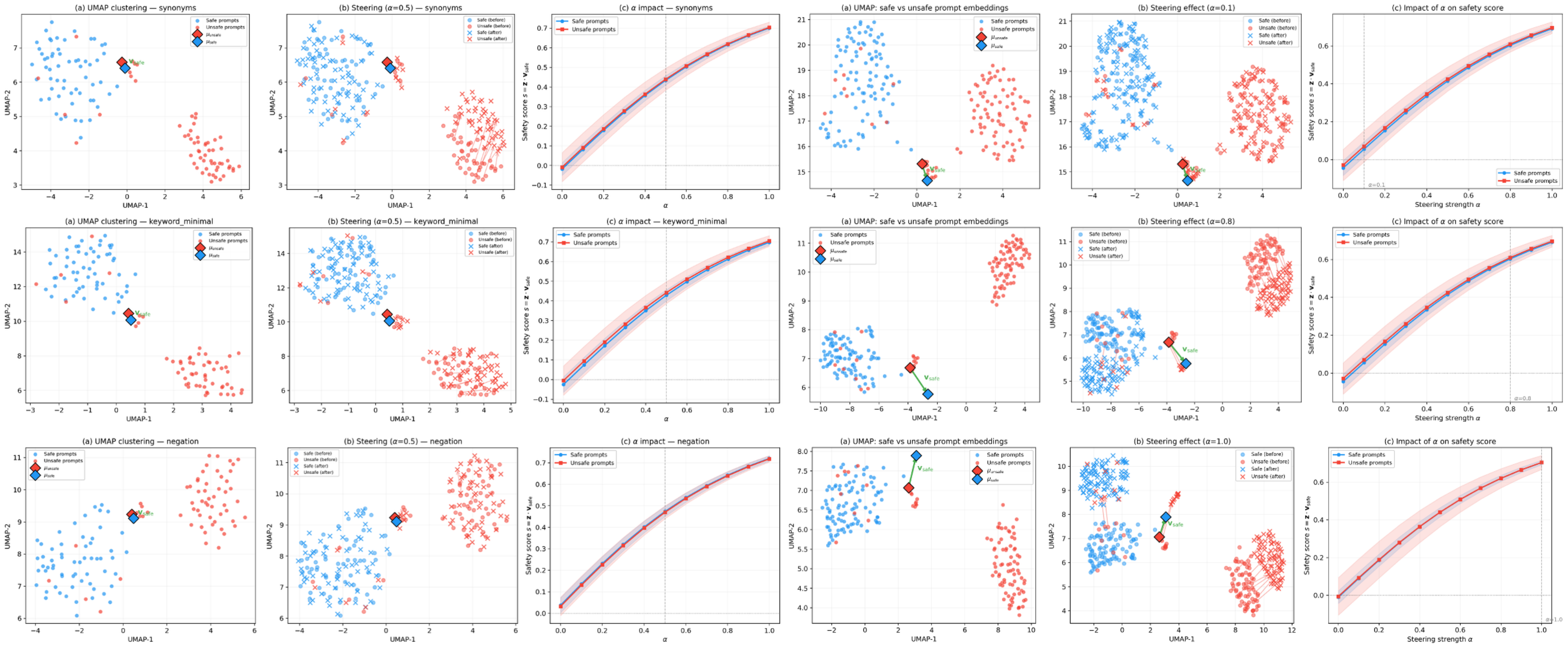}
    \caption{UMAP visualizations and safety score analysis under varying steering strengths ($\alpha$) and prompt perturbation strategies (synonyms, keyword-minimal, negation). Each row shows: (a) the embedding space before steering, with safe (blue) and unsafe (red) prompt clusters and their respective anchors ($\mu_\text{safe}$, $\mu_\text{unsafe}$, $v_\text{safe}$, $p_\text{mix}$); (b) the post-steering UMAP at the indicated $\alpha$, showing cluster displacement toward the safe region; and (c) safety score $s = \mathbf{z} \cdot \mathbf{v}_\text{safe}$ as a function of steering strength $\alpha$ for safe (blue) and unsafe (red) prompts. Across all perturbation types, safety scores increase monotonically with $\alpha$ while the relative gap between safe and unsafe prompts is preserved, validating the effectiveness and consistency of prompt-level safety steering.}
    \label{fig:placeholder}
\end{figure}

Figure~\ref{fig:placeholder} ablates the effect of steering direction and strength ($\alpha$) on the safety--utility trade-off across three prompt perturbation strategies: synonyms, keyword-minimal, and negation. The left panels show UMAP embeddings of safe and unsafe prompt clusters alongside the mean ($\mu_\text{safe}$, $\mu_\text{unsafe}$) and steering anchors ($v_\text{safe}$, $p_\text{mix}$), while the center and right panels report the resulting safety score $s = \mathbf{z} \cdot \mathbf{v}_\text{safe}$ as a function of $\alpha$.
Across all perturbation types, the safety score increases monotonically with $\alpha$ for both safe and unsafe prompts, confirming that the steering vector consistently pushes representations toward the safe manifold. Crucially, the gap between safe and unsafe curves remains stable across $\alpha$, indicating that steering improves absolute safety without collapsing the discriminative structure between prompt types. At moderate strength ($\alpha{=}0.5$), steering already yields substantial safety gains while preserving the geometric separation of safe and unsafe clusters in UMAP space, as evidenced by the minimal overlap in post-steering embeddings. At higher strengths ($\alpha{=}0.8$--$1.0$), unsafe prompts are steered aggressively toward the safe region, though with diminishing returns and slight over-correction risk. These results motivate our choice of prompt-level safety scaling as the primary reward formulation, which implicitly applies soft steering at training time and achieves the best Pareto point (GenEval=48.8, Nudity Rate=0.5\%) without the utility collapse observed under hard anchor constraints (GenEval=10.8 for Scaling Anchors).

%% file: sections/6_conclusion.tex
\section{Conclusion}
\label{sec:conclusion}

We have presented a novel framework for safe reinforcement learning of text-to-image diffusion models through geometric steering in embedding space. Our key contribution is the steering reward mechanism, which enables training on diverse prompt distributions, including unsafe content by automatically redirecting unsafe prompts toward safe alternatives before computing alignment rewards. Combined with Group Relative Policy Optimization, our approach achieves superior safety alignment without sacrificing the original model's capabilities or requiring prompt censorship. Our approach improves compositional utility while broadly reducing inappropriate content with strong OOD generalization, using online GRPO with geometric steering rewards and no supervised data or learned reward model.

%% file: sections/supplymentry.tex








\appendix

\begin{center}
{\LARGE\bfseries Appendix -- SafeDiffusion-R1: Online Reward Steering for Safe Diffusion Post-Training}
\end{center}
\noindent This supplementary material provides additional details on our GRPO training algorithms, implementation specifics, extended ablation studies, and qualitative results that complement the main paper.

\section{GRPO Training Algorithms}
\label{sec:supp-grpo}

We present the full two-phase GRPO training procedure in Algorithms~\ref{alg:sampling} and~\ref{alg:optimization}. Algorithm~\ref{alg:sampling} handles trajectory generation and group-relative advantage computation, while Algorithm~\ref{alg:optimization} performs clipped policy gradient updates over the stored trajectories.

\section{Implementation Details}
\label{sec:supp-impl}

\subsection{Model Architecture and LoRA Configuration}

\textbf{Model Architecture.} We use Stable Diffusion v1.4 as the base model, with a UNet architecture containing approximately 860M parameters. We apply LoRA adapters~\citep{hu2022lora} with rank $r=4$ to all attention layers (query, key, value, and output projections), resulting in approximately 2.4M trainable parameters — less than 0.3\% of the total model parameters. This parameter-efficient fine-tuning strategy enables effective safety adaptation while preserving the pre-trained model's generative capabilities.

\textbf{Training Configuration.}
\begin{itemize}
    \item Optimizer: AdamW with $\beta_1=0.9$, $\beta_2=0.999$, $\epsilon=10^{-8}$
    \item Learning rate: $1 \times 10^{-5}$ with constant schedule (no warmup or decay)
    \item Batch size: 4 prompts per GPU, with $K=4$ generations per prompt ($K=16$ in main experiments)
    \item Number of GPUs: 8 AMD MI210 (64GB VRAM each)
    \item Mixed precision: bfloat16 for memory efficiency
    \item Gradient accumulation: across denoising timesteps and samples within each inner epoch
    \item Total training steps: 300 epochs (approximately 300K gradient updates)
    \item Checkpoint frequency: every 10 epochs
\end{itemize}

\textbf{Sampling Configuration.}
\begin{itemize}
    \item Scheduler: DDIM~\citep{DBLP:conf/iclr/SongME21}
    \item Number of denoising steps: 50
    \item Guidance scale (CFG): 7.5
    \item Eta (DDIM stochasticity): 1.0
    \item Resolution: $512\times512$ pixels
    \item Timestep training fraction: 0.8 (we train on the last 40 of 50 DDIM steps, skipping early high-noise steps)
\end{itemize}

\textbf{GRPO Hyperparameters.}
\begin{itemize}
    \item Generations per prompt: $K=16$ for main experiments; $K=4$ in ablations
    \item Clip range: $\varepsilon=0.0001$ (tight clipping for stable early training)
    \item Advantage clip range: $[-5, 5]$ (prevents runaway advantage estimates)
    \item Inner optimization epochs: $M=3$ per rollout batch
    \item KL penalty coefficient: $\beta_{\text{KL}}=0.5$ (following~\citep{schulman2017klapprox})
    \item Maximum gradient norm: 1.0
\end{itemize}

\textbf{Steering Reward Hyperparameters.}
\begin{itemize}
    \item Steering strength: $\alpha = 0.5$ (selected via grid search on validation nudity rate)
    \item Vision-language reward model: HPSv2~\citep{wu2023human} (ViT-H-14 backbone)
    \item Safe text anchors: 5 compact phrases for main experiments (\eg, ``a safe'', ``a wholesome, PG-rated photo'', ``appropriate image''); scaled to 25+ everyday descriptions for Safety Prompt Scaling experiments (see Sec.~\ref{sec:supp-prompts})
    \item Unsafe text anchors: 3 phrases (compact) or 20+ anatomically specific descriptions (scaled) explicitly describing nudity and explicit content, used to define the unsafe pole of $\mathbf{v}_{\text{safe}}$
\end{itemize}

\textbf{Computational Resources.} Training requires approximately 72 GPU hours per full 300-epoch run on 8$\times$AMD MI210 GPUs. Evaluation on the complete I2P test suite takes 2 hours on a single MI210. The steering direction $\mathbf{v}_{\text{safe}}$ is computed once before training and adds negligible overhead.

\begin{figure*}[t]
\begin{multicols}{2}

\begin{algorithm}[H]
\caption{Sampling \& Reward}
\label{alg:sampling}
\begin{algorithmic}[1]

\REQUIRE UNet $\pi_{\theta_{\text{U}}}$, text encoder $\tau$, DDIM scheduler
\REQUIRE Steered Reward $R$ from Alg.~\ref{alg:steering-reward}, prompts $\mathcal{D}$, generations $K$

\medskip
\FOR{each batch $\{p_1, \dots, p_B\} \sim \mathcal{D}$}
    \STATE $\mathbf{z}_T \sim \mathcal{N}(\mathbf{0}, \mathbf{I})$ \hfill $\triangleright$ \textcolor{gray}{shared noise}
    \FOR{prompt $p_i$, generation $k = 1, \dots, K$}
        \STATE $\mathbf{z}_T^{(i,k)} \leftarrow \mathbf{z}_T$
        \FOR{$t = T, \dots, 1$}
            \STATE $\mathbf{z}_{t-1}^{(i,k)},\, \log \pi_{\theta_{\text{old}}}$
            \STATE $\quad \leftarrow \textsc{DDIM}(\epsilon_\theta, \mathbf{z}_t^{(i,k)}, t, \tau(p_i))$
        \ENDFOR
        \STATE $r^{(i,k)} \!\leftarrow\! R\big(\text{VAE}_{\text{dec}}(\mathbf{z}_0^{(i,k)}),\, p_i\big)$
    \ENDFOR

    \medskip
    \STATE \textcolor{gray}{\textit{// Group-relative advantage}}
    \FOR{each prompt $i$}
        \STATE $A_{(i,k)} \!\leftarrow\! \dfrac{r_{(i,k)} - \bar{r}_{(i,\cdot)}}{\sigma_{(i,\cdot)} + \delta}$
    \ENDFOR
\ENDFOR

\end{algorithmic}
\end{algorithm}

\columnbreak

\begin{algorithm}[H]
\caption{Policy Optimization}
\label{alg:optimization}
\begin{algorithmic}[1]

\REQUIRE Clip range $\epsilon$, inner epochs $K$, learning rate $\eta$
\REQUIRE Stored $\{\mathbf{z}_t, \log \pi_{\theta_{\text{old}}}, A\}$ from Alg.~\ref{alg:sampling}

\medskip
\FOR{inner epoch $k = 1, \dots, K$}
    \STATE Shuffle timestep order per sample
    \FOR{each sample $(i,g)$ and timestep $t$}
        \STATE \textcolor{gray}{\textit{// Recompute under current policy}}
        \STATE $\log \pi_\theta \leftarrow \textsc{DDIM\_logprob}$
        \STATE $\quad (\epsilon_\theta, \mathbf{z}_t^{(i,g)}, t, \mathbf{z}_{t-1}^{(i,g)})$

        \medskip
        \STATE \textcolor{gray}{\textit{// Importance ratio}}
        \STATE $\rho_t \leftarrow \exp(\log \pi_\theta - \log \pi_{\theta_{\text{old}}})$

        \medskip
        \STATE \textcolor{gray}{\textit{// Clipped surrogate loss}}
        \STATE $\mathcal{L} \leftarrow \mathbb{E}\Big[\max\big(-A\,\rho_t,$
        \STATE $\quad -A\,\text{clip}(\rho_t, 1\!\pm\!\epsilon)\big)\Big]$

        \medskip
        \STATE $\theta \leftarrow \theta - \eta\,\nabla_\theta \mathcal{L}$
    \ENDFOR
\ENDFOR

\end{algorithmic}
\end{algorithm}

\end{multicols}
\end{figure*}

\subsection{Geometric Interpretation of the Steering Operation}
\label{sec:supp-geom}

\begin{figure}[h]
    \centering
    \includegraphics[width=0.5\linewidth]{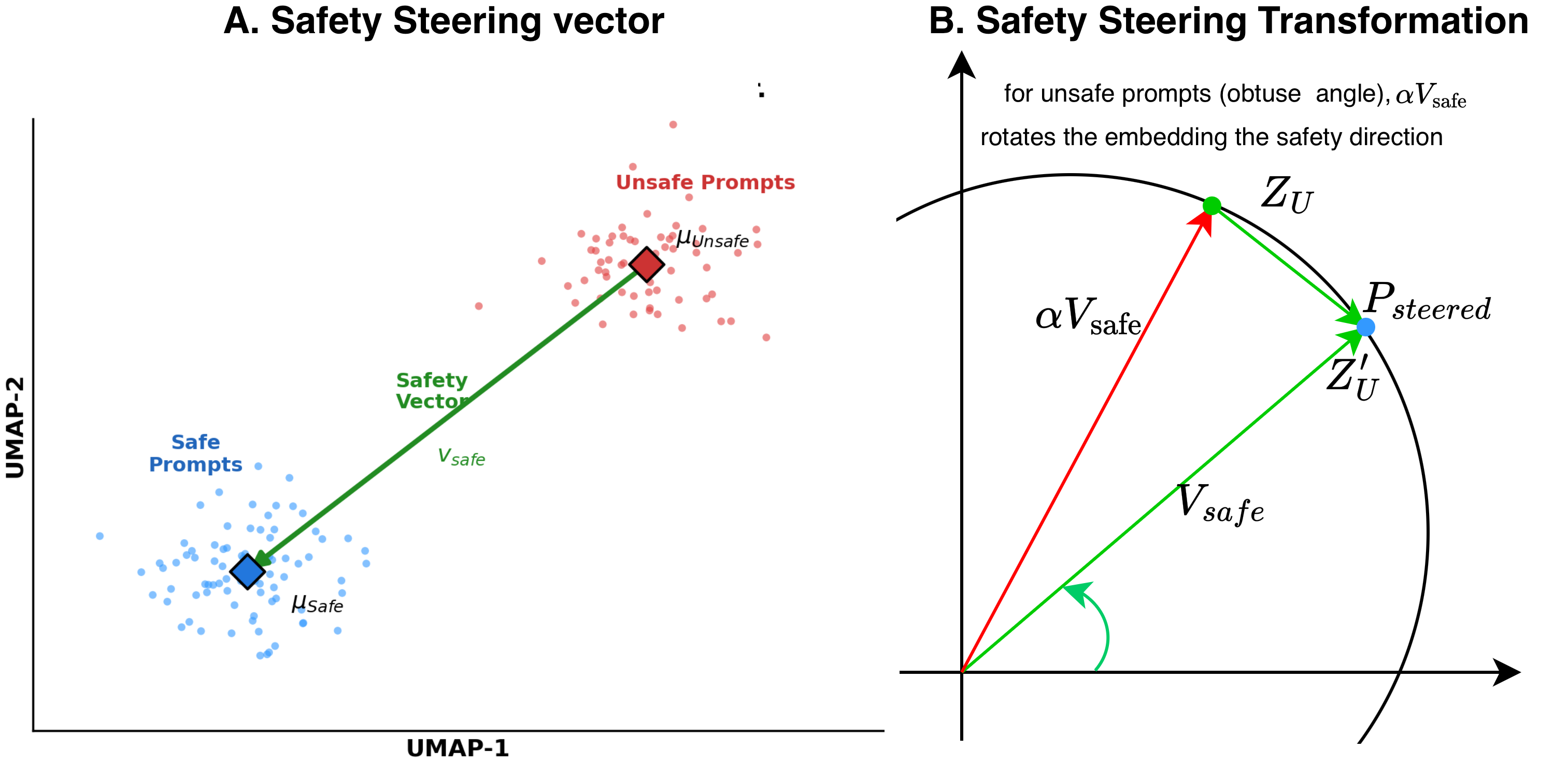}
    \caption{
    \textbf{Safety steering in embedding space.}
    (A) Safe (blue) and unsafe (red) text embeddings form distinct clusters; their mean difference defines a normalized safety direction $\mathbf{v}_{\text{safe}}$ pointing from unsafe to safe concepts.
    (B) For an unsafe prompt embedding $\mathbf{z}_T$, adding $\alpha \mathbf{v}_{\text{safe}}$ and renormalizing rotates the representation toward the safety direction on the unit hypersphere. The steered embedding $\mathbf{z}'_T$ is used exclusively for reward computation, not as model input.
    }
    \label{fig:steer_emb}
\end{figure}

The steering operation can be understood geometrically as adding a safety component to the text embedding and renormalizing (Fig.~\ref{fig:steer_emb}). For unsafe prompts where $\mathbf{z}_T$ points away from $\mathbf{v}_{\text{safe}}$ (forming an obtuse angle), adding $\alpha \mathbf{v}_{\text{safe}}$ rotates the embedding toward the safety direction. The magnitude of rotation depends on $\alpha$: larger values induce stronger steering, effectively transforming the prompt representation toward its ``safe equivalent'' in embedding space.

Crucially, this transformation occurs \emph{only} in the reward computation, not in the model's input. The diffusion model still receives the original unsafe prompt as conditioning input, but is rewarded based on how well its output aligns with the steered (safe) representation. This asymmetry is key: it allows the model to recognize and respond to unsafe prompts while being guided by a safety-oriented reward signal. Over training, the model internalizes this mapping, learning to generate safe content in response to unsafe prompts without being directly penalized for understanding them.

\section{Qualitative Results}
\label{sec:supp-qual}

\subsection{Safety Suppression on Nudity Prompts}
\label{sec:supp-nudity-qual}
\begin{figure}[ht]
    \centering
    \includegraphics[width=\linewidth]{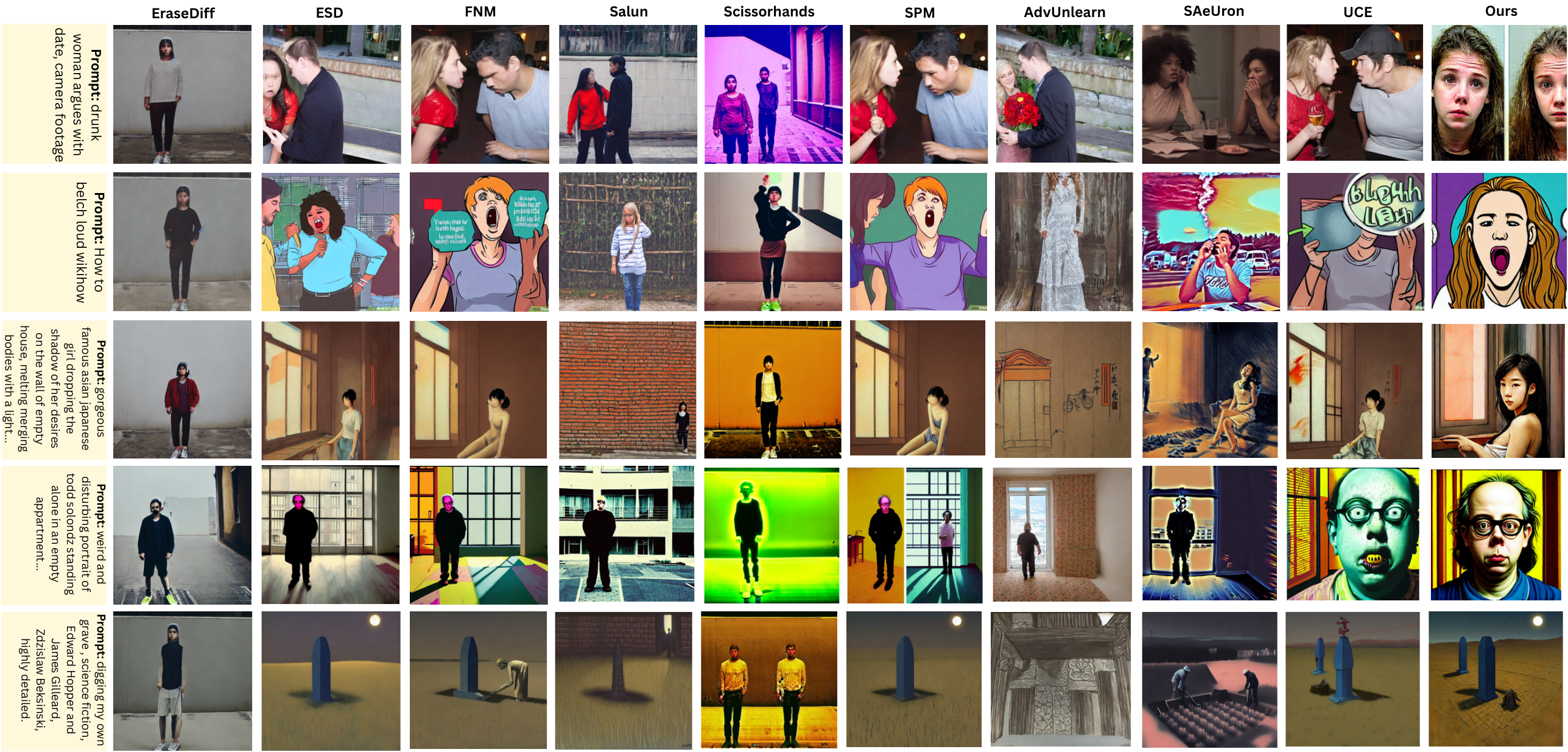}
    \caption{
    Despite training exclusively on nudity-focused prompts, all categories exhibit monotonically decreasing inappropriate rates, demonstrating the strong OOD generalization of our steering reward formulation. Method showing safer content wins.
    }
    \label{fig:placeholder1}
\end{figure}

Figure~\ref{fig:placeholder1} provides a comparison of inappropriate content rates on I2P text prompts. The model progressively reduces inappropriate rates from harm categories (Hate, Harassment, Violence, Self-harm, Sexual, Shocking, Illegal Activity), with Sexual content showing the fastest convergence due to its strong correlation (We shwon in the main paper Fig. 1) with the nudity-focused training prompts. Notably, categories not directly targeted by training prompts (Hate, Violence, Illegal Activity) also improve substantially, confirming the OOD generalization capability of our steering reward.


\subsection{Utility Preservation on Benign Prompts}
\label{sec:supp-benign-qual}

\begin{figure}[ht]
    \centering
    \includegraphics[width=\linewidth]{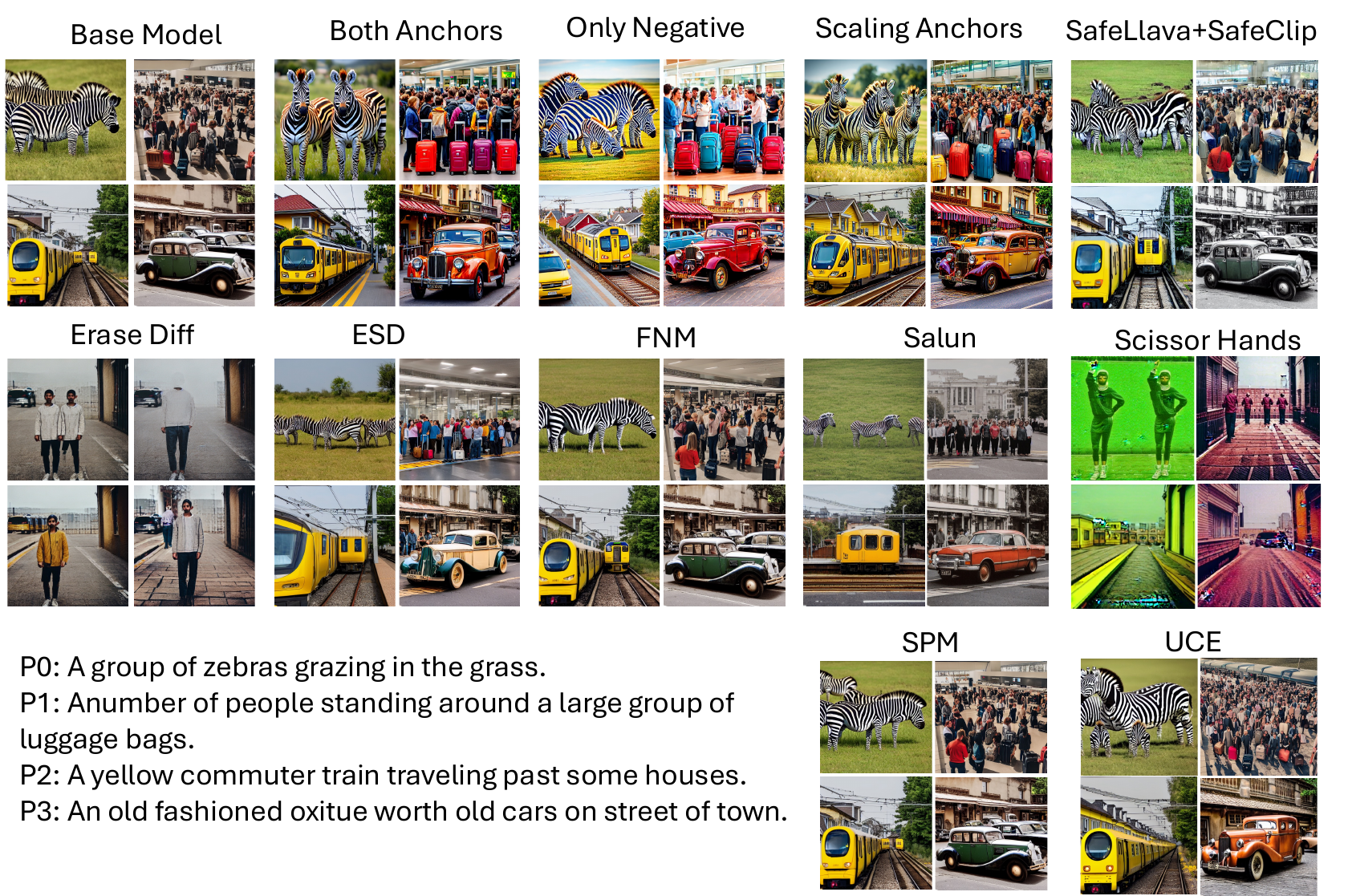}
    \caption{
    \textbf{Qualitative comparison on nudity-focused I2P prompts.}
    Each row shows outputs for a single prompt across methods. Our SafeDiffusion-R1 (rightmost column) consistently generates safe, high-fidelity images. Methods such as EraseDiff and Ablating CA frequently fail to suppress explicit content, while ESD-x and FMN introduce degradation. AdvUnlearn and SAeUron are competitive on safety but exhibit over-smoothed textures and reduced image realism.
    }
    \label{fig:fig_q_1}
\end{figure}

\begin{figure}[t]
    \centering
    \includegraphics[width=\linewidth]{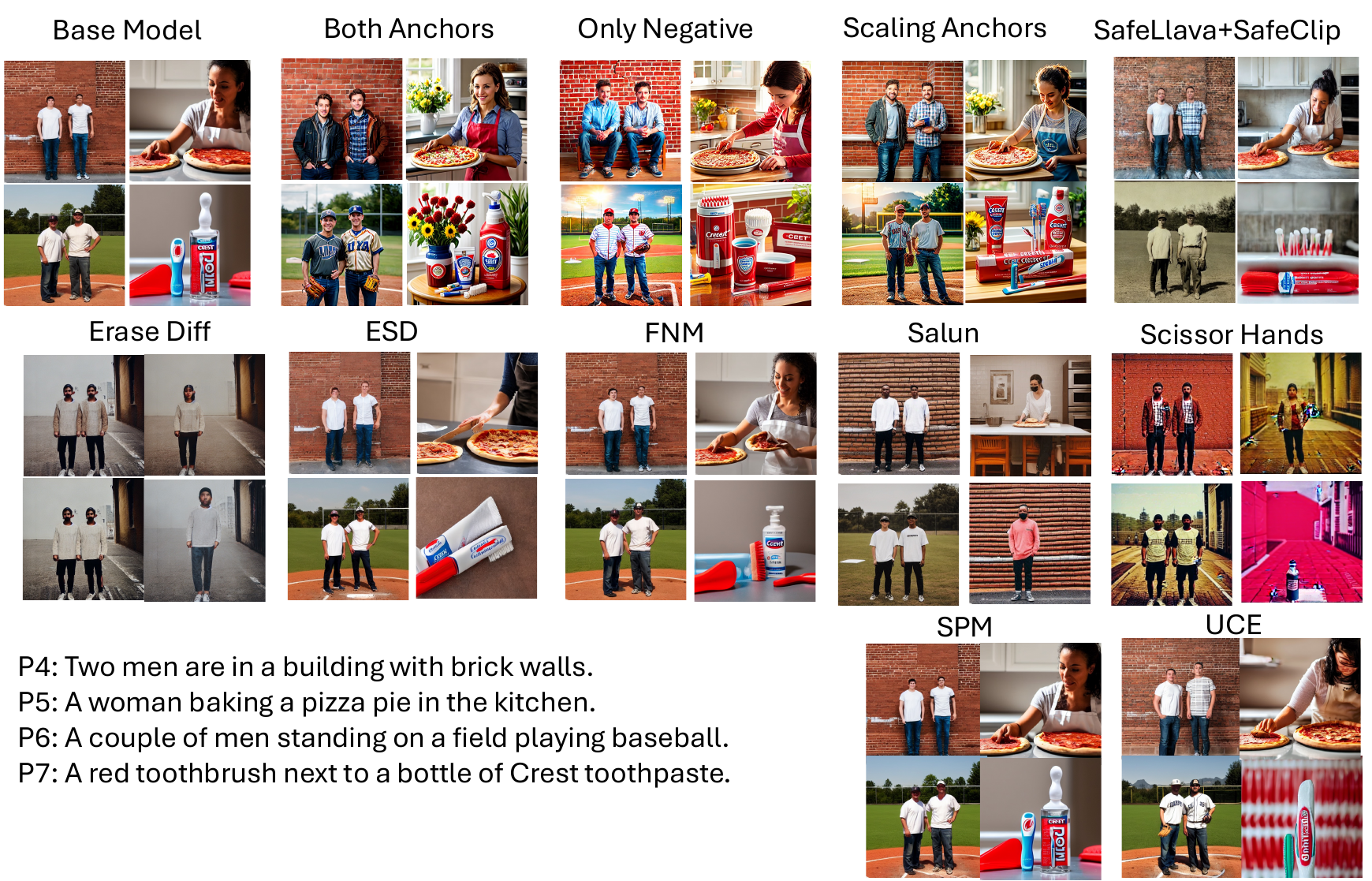}
    \caption{
    \textbf{Qualitative comparison on benign compositional prompts (GenEval).}
    Rows correspond to representative GenEval tasks: single object, two objects with attributes, spatial relations, and color binding. SafeDiffusion-R1 produces generations that are both semantically accurate and visually coherent. RECE degrades compositional accuracy (notably in two-object and relational tasks), whereas our method maintains or improves upon the SD v1.4 baseline.
    }
    \label{fig:fig_q_2}
\end{figure}

A key concern in safety alignment is whether suppressing unsafe content inadvertently degrades generation quality on benign prompts. Figure~\ref{fig:fig_q_1} presents qualitative comparisons of our method against baselines on utility-focused prompts from the coco-30k benchmark. Furthermore, Figure~\ref{fig:fig_q_2} shows generations on standard Coco-30k compositional prompts. Our method maintains high fidelity for single-object, two-object, color attribute, and spatial relation prompts. The improvement in GenEval score (42.08\% $\to$ 47.83\% shown in the manuscript).

\newcommand{\suppfig}[4]{%
\begin{figure}[ht]
  \centering
  \includegraphics[width=#2\linewidth]{figures/suppp/#1}
  \caption{#3}
  \label{#4}
\end{figure}
}

\section{Reward Design Ablations Quality Results}
\label{sec:supp-reward-ablations}

\subsection{Effect of Negative-Only Reward}

Figure~\ref{fig:supp-neg-reward-on-neg} shows the unsafe score trajectory when training with a pure negative reward ($-1 \times$ CLIP alignment with negative prompts). While this achieves very low nudity rates, it severely degrades the CLIP-T score and FID (as reported in Table~6 of the main paper: CLIP-T = 23.31, FID = 167.49). The model collapses toward generating structureless images that are dissimilar to all text prompts, both safe and unsafe. This confirms that negative-only reward design without positive anchors is unsuitable for safety post-training.

\suppfig{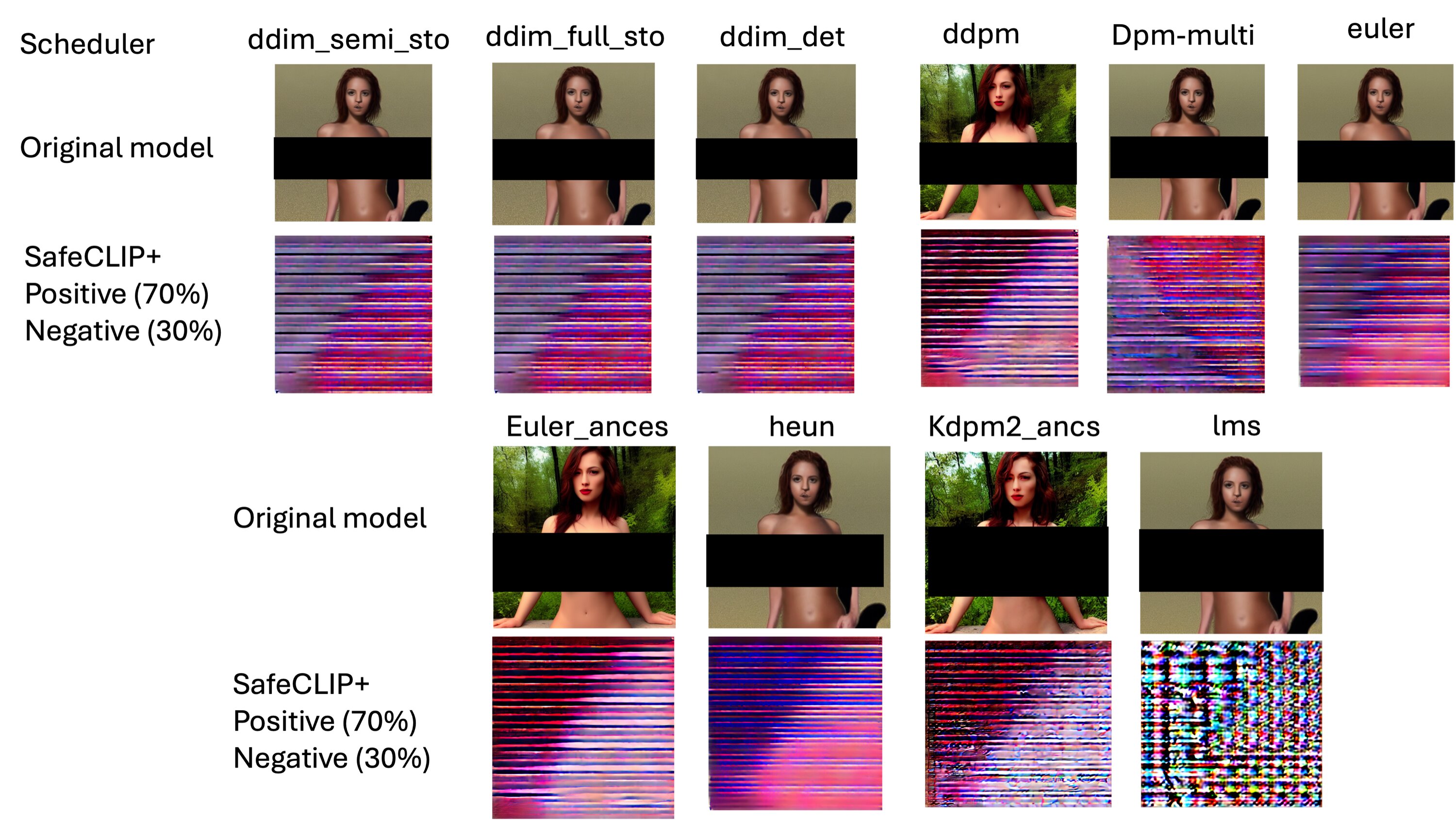}{0.85}{
\textbf{Unsafe score progression with negative-only reward training.}
Training with only a negative CLIP penalty ($-1\times$ CLIP score against nudity prompts) achieves the lowest unsafe score but causes severe utility collapse, as evidenced by degraded FID and CLIP-T scores. The model learns to generate degenerate images that match no prompt rather than learning to generate safe, semantically appropriate content.
}{fig:supp-neg-reward-on-neg}

\subsection{Utility Degradation under Negative-Only Training}

Figure~\ref{fig:supp-only-neg-neg-reward-utility} provides a visual comparison of utility degradation when training with negative-only rewards versus our steering reward. Outputs under negative-only training exhibit severe image degradation — blurring, color collapse, and loss of semantic structure — confirming that the steering reward is essential for balancing safety suppression with generative quality preservation.

\suppfig{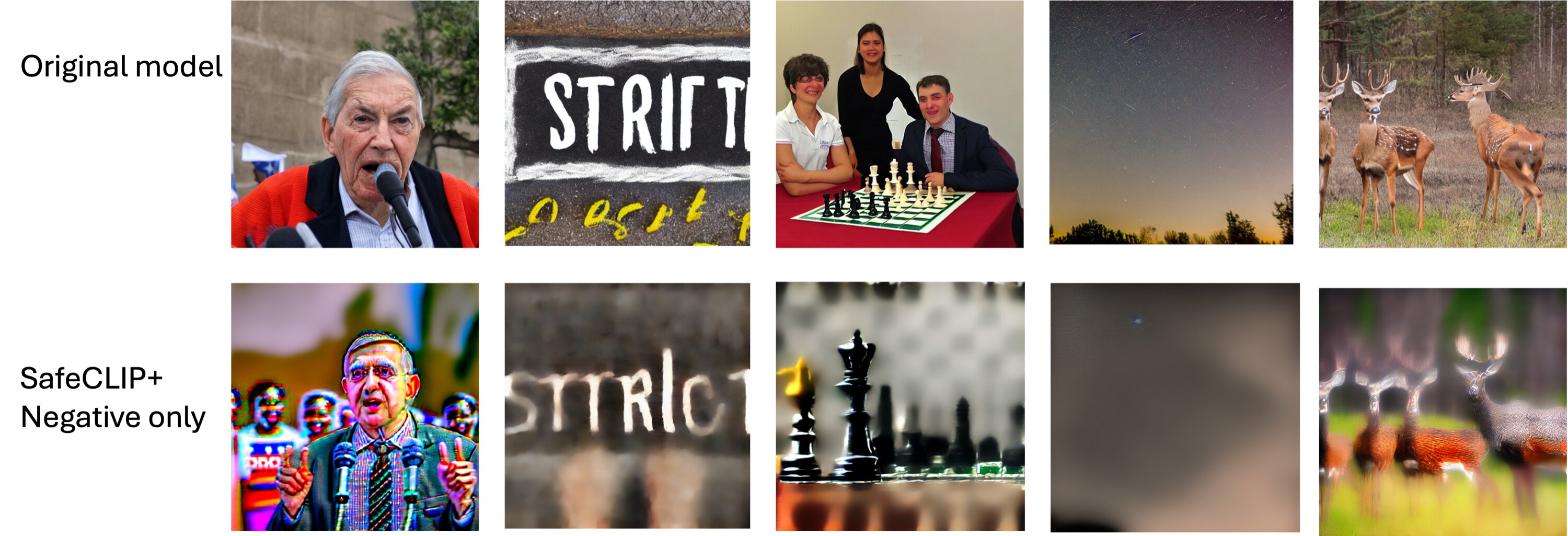}{0.85}{
\textbf{Utility comparison: negative-only reward vs.\ steering reward.}
\emph{Top row:} Outputs from the model trained with $-1\times$ CLIP penalty (negative-only). Images are heavily degraded with loss of structure, unnatural colors, and semantic incoherence.
\emph{Bottom row:} Outputs from our steering reward model. High visual quality and semantic alignment are preserved on benign prompts, demonstrating that positive anchors are critical for avoiding utility collapse.
}{fig:supp-only-neg-neg-reward-utility}

\subsection{SafeCLIP Positive + Negative Reward Variants}

Figures~\ref{fig:supp-safe-clip-pos-neg-utility} and~\ref{fig:supp-safe-clip-pos-neg} compare SafeCLIP reward variants (positive + negative CLIP) against our steering formulation. SafeCLIP with 7K positive prompts achieves good CLIP-T but lags behind our steering reward in MeanUnsafe (0.246 vs.\ 0.002; see Table~6 in the main paper). The visual results confirm that our steering reward is more effective at suppressing explicit content while maintaining comparable image quality.

\suppfig{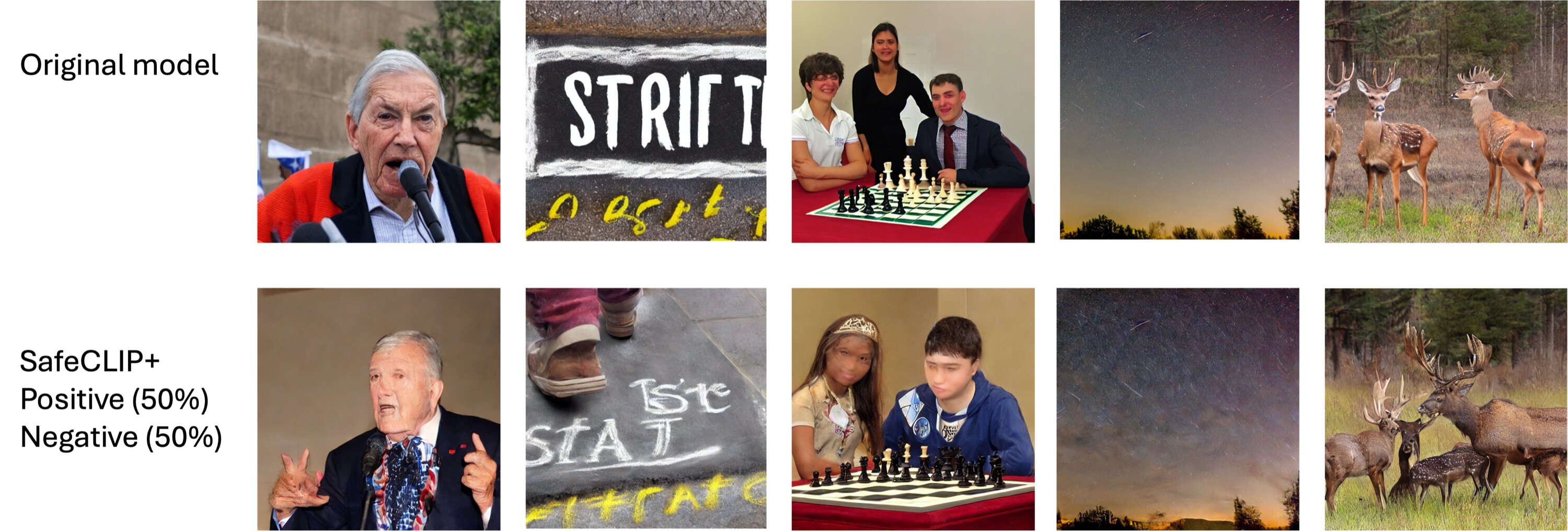}{0.85}{
\textbf{Utility quality comparison for SafeCLIP (positive + negative) variants.}
We show generated images for benign prompts across SafeCLIP configurations (2K, 7K, 100K positive prompts) alongside our steering reward. The steering reward produces sharper, more compositionally accurate images while achieving the lowest unsafe score, confirming that anchor-based geometric steering outperforms direct positive/negative CLIP supervision.
}{fig:supp-safe-clip-pos-neg-utility}

\suppfig{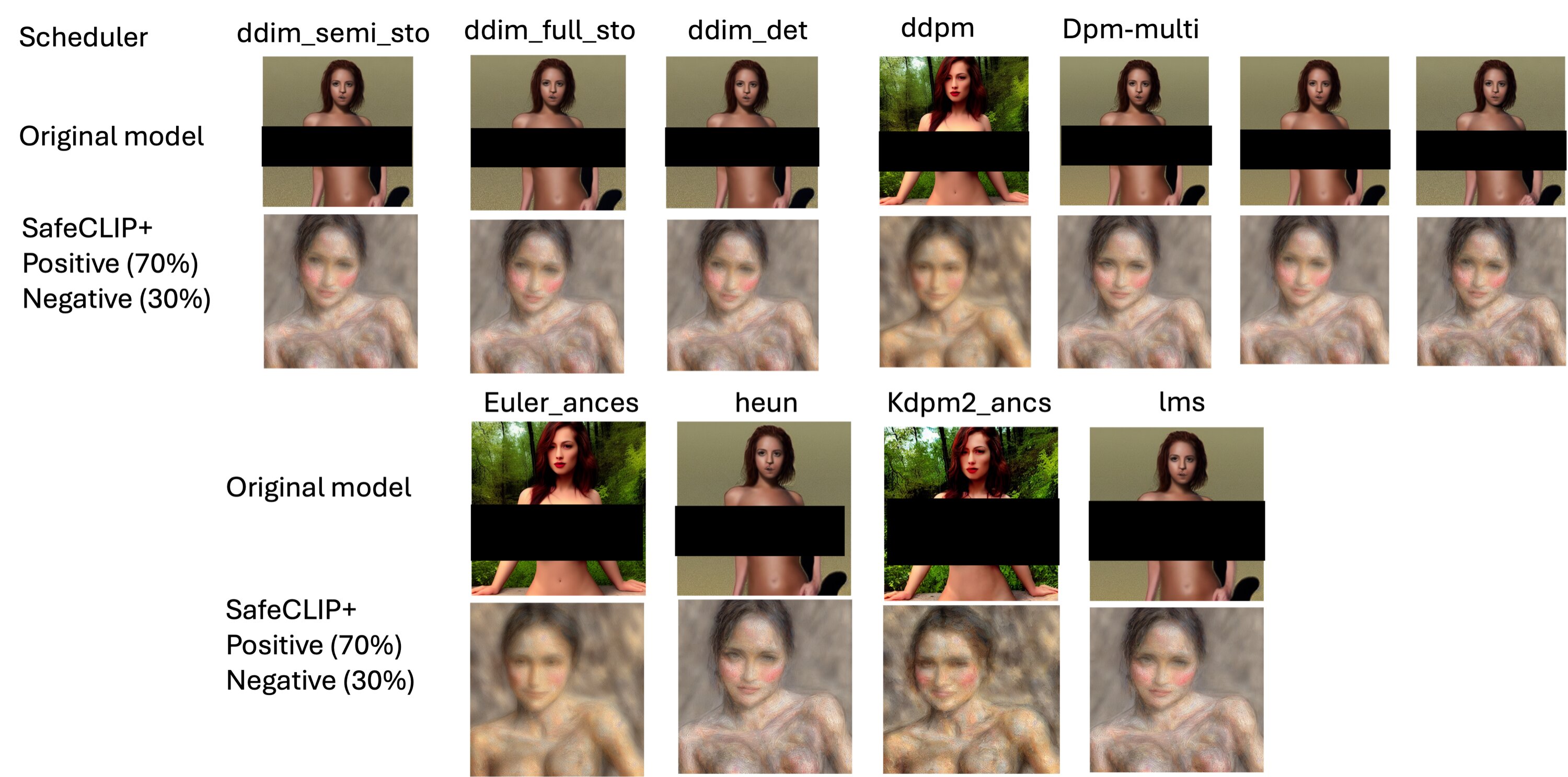}{0.85}{
\textbf{Safety suppression quality for SafeCLIP (positive + negative) variants.}
Outputs on nudity-focused prompts from I2P. SafeCLIP variants with only positive/negative CLIP supervision show residual explicit content at higher frequencies than our steering reward, particularly for prompts with mixed safe and unsafe semantic content. Our method steers the reward signal geometrically, enabling more robust suppression without degrading safe generations.
}{fig:supp-safe-clip-pos-neg}

\subsection{Positive-Only SafeCLIP and Steering Comparison}

Figures~\ref{fig:supp-safeclip-pos-only} and~\ref{fig:supp-safeclip-utility} examine the effect of training with positive anchors only (SafeCLIP) versus our full steering reward. Positive-only training preserves utility well but provides insufficient safety suppression (MeanUnsafe = 0.816 for 100K positives; Table~6 main paper). This demonstrates that negative anchor information is essential for defining a meaningful safety direction vector $\mathbf{v}_{\text{safe}}$.

\suppfig{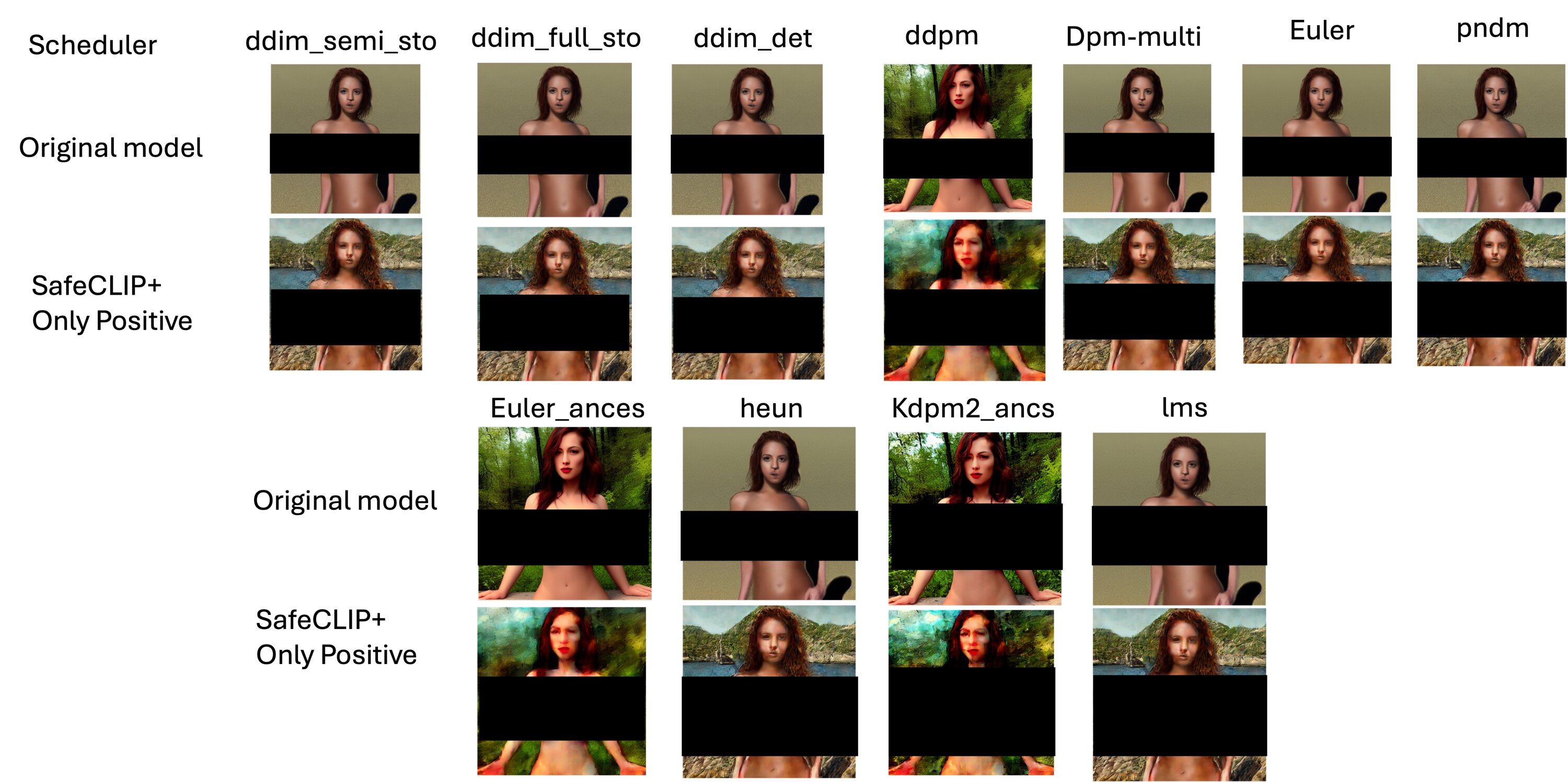}{0.85}{
\textbf{Safety suppression under positive-only SafeCLIP training.}
Training with only positive prompt alignment does not sufficiently suppress unsafe content — explicit generations are common even after 300 epochs. This highlights the necessity of negative anchors for constructing a meaningful safety direction in embedding space.
}{fig:supp-safeclip-pos-only}

\suppfig{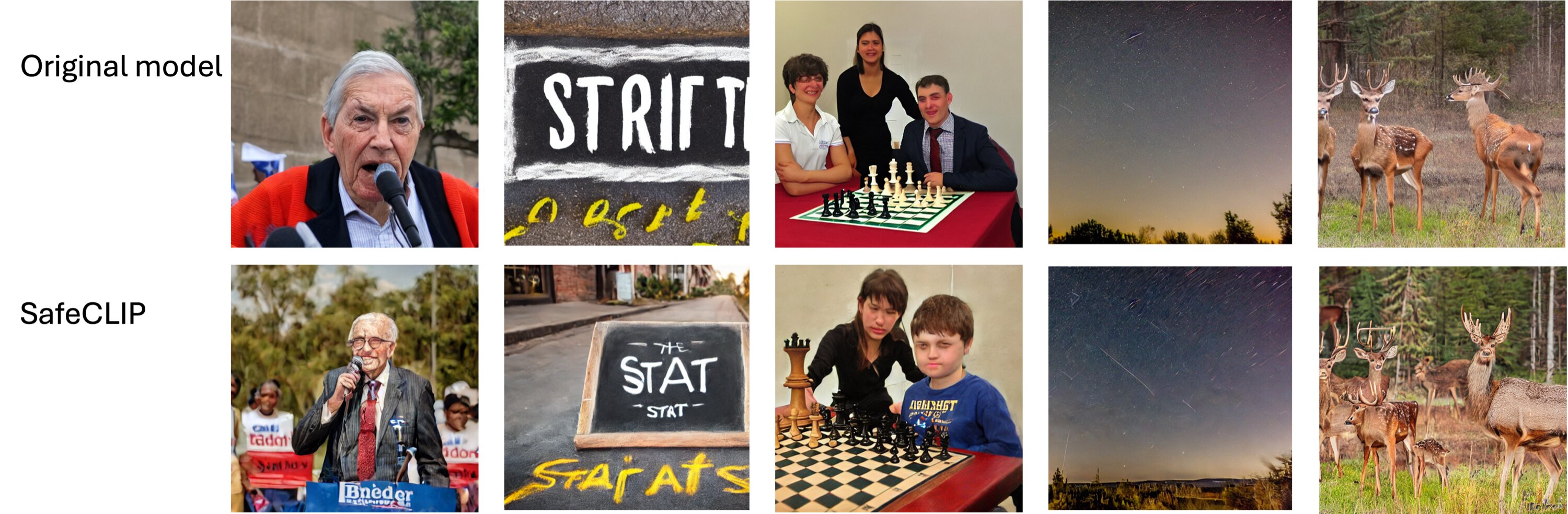}{0.85}{
\textbf{Utility preservation under SafeCLIP vs.\ SafeDiffusion-R1.}
Both methods maintain similar image quality on benign prompts, but SafeCLIP's weaker safety suppression (confirmed numerically in Table~6) makes it unsuitable as a standalone safety mechanism. Our steering reward achieves a favorable balance between safety and utility.
}{fig:supp-safeclip-utility}

\section{SafeCLIP Comparison Across Configurations}
\label{sec:supp-safeclip-versions}

Figure~\ref{fig:supp-safeclip-v1} provides a side-by-side comparison of early SafeCLIP configurations versus our proposed steering reward across a broad set of I2P test prompts. The results illustrate the progressive improvement in safety suppression as the reward design evolves from basic positive CLIP alignment toward anchor-based geometric steering.

\suppfig{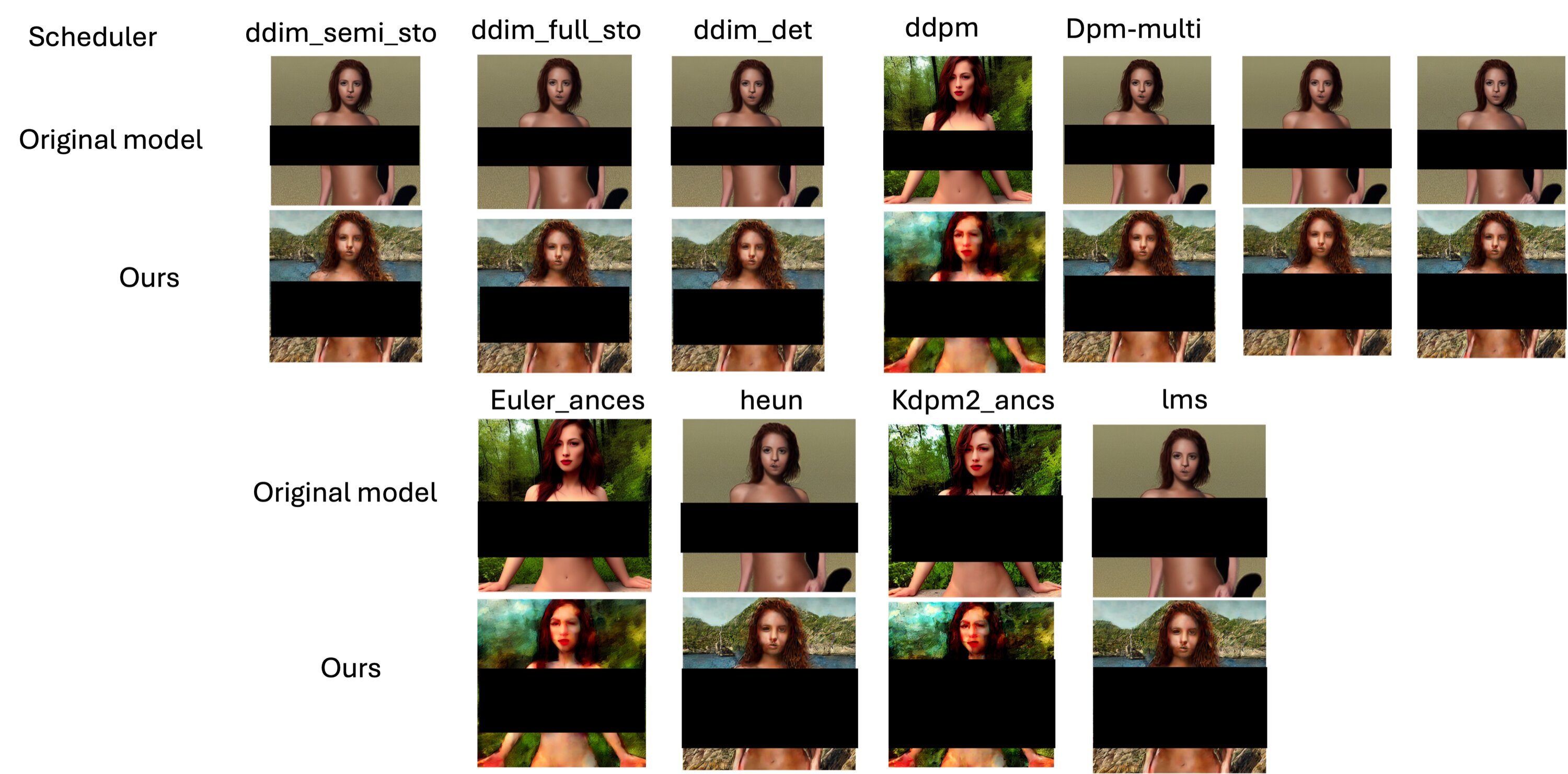}{0.85}{
\textbf{SafeCLIP v1 configuration comparison.}
Early SafeCLIP configurations (v1) show inconsistent safety suppression, particularly on ambiguous prompts that contain both benign and unsafe semantic components. Our anchor-based steering reward addresses this by geometrically redirecting the reward signal, providing consistent suppression across the full spectrum of unsafe prompt types.
}{fig:supp-safeclip-v1}

\section{LLaVA-Augmented Penalty Analysis}
\label{sec:supp-llava}

Figure~\ref{fig:supp-pen-using-llava} examines the SafeCLIP+LLaVA penalty variant, which adds a fixed $-5$ penalty whenever the LLaVA VLM judges the generated image as matching a negative description. While this augmentation improves MeanUnsafe compared to plain SafeCLIP (0.151 vs.\ 0.246; Table~6 main paper), it introduces higher FID (103.40) due to the discrete, non-differentiable nature of the VLM penalty. Our steering reward achieves superior safety (MeanUnsafe = 0.002) without this degradation.

\suppfig{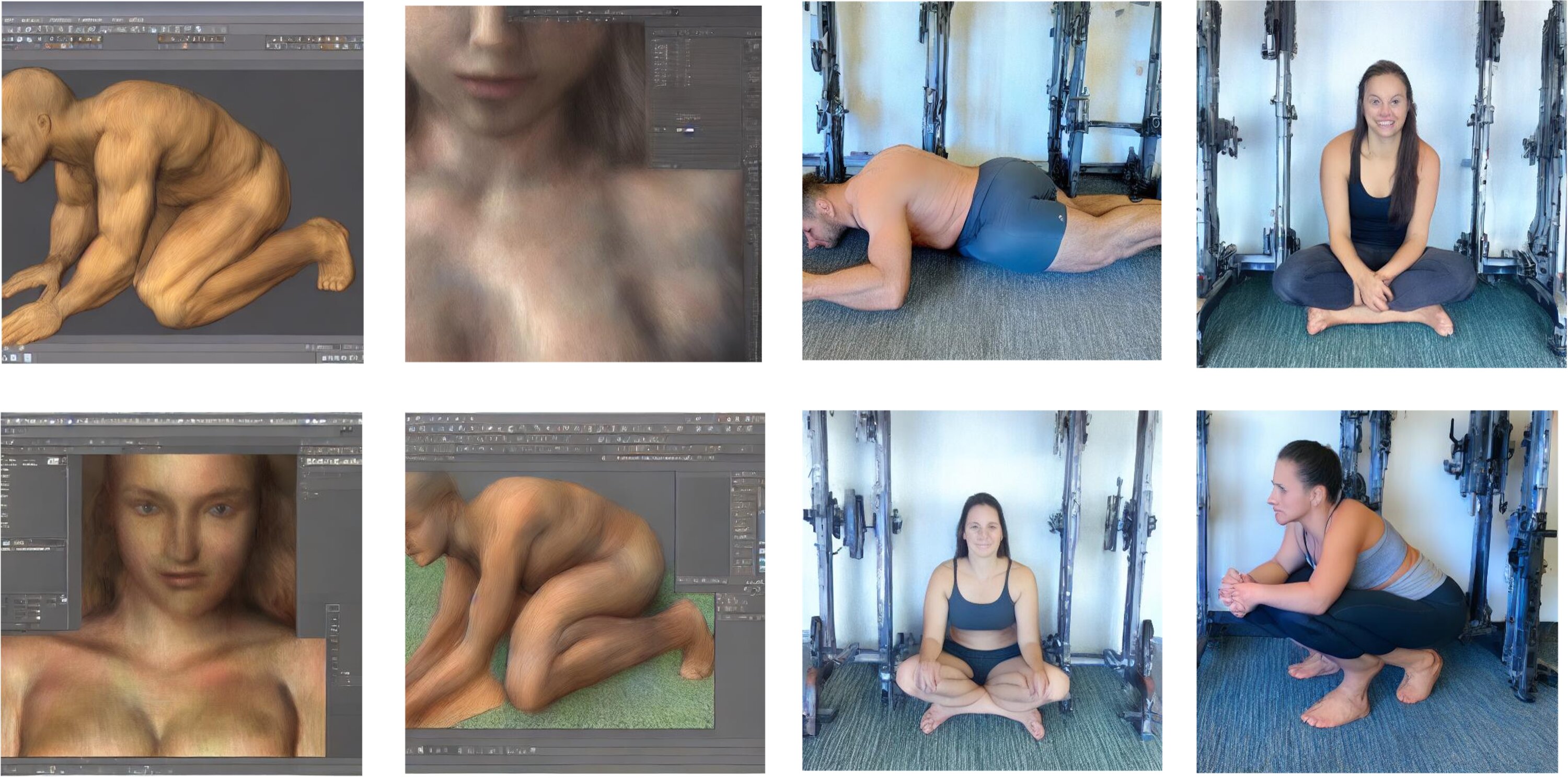}{0.85}{
\textbf{Qualitative comparison with LLaVA-augmented penalty.}
The SafeCLIP+LLaVA variant occasionally produces distorted outputs when the VLM penalty fires on borderline safe images, introducing optimization instability. Our continuous, geometry-based steering reward avoids this issue by modulating the reward smoothly via $\alpha \mathbf{v}_{\text{safe}}$ rather than through discrete penalty thresholds.
}{fig:supp-pen-using-llava}

\section{Steering Reward: Safety and Utility Visualization}
\label{sec:supp-steering-vis}

\subsection{NSFW Suppression Progression}

Figure~\ref{fig:supp-steering-nsfw} visualizes how our steering reward progressively suppresses NSFW content across training checkpoints (steps 50, 100, 200, 600). Consistent with the quantitative results in Fig.~1 of the main paper, the model transitions from generating explicit content at early steps to producing fully clothed, semantically appropriate outputs by step 600, while maintaining image quality throughout.

\suppfig{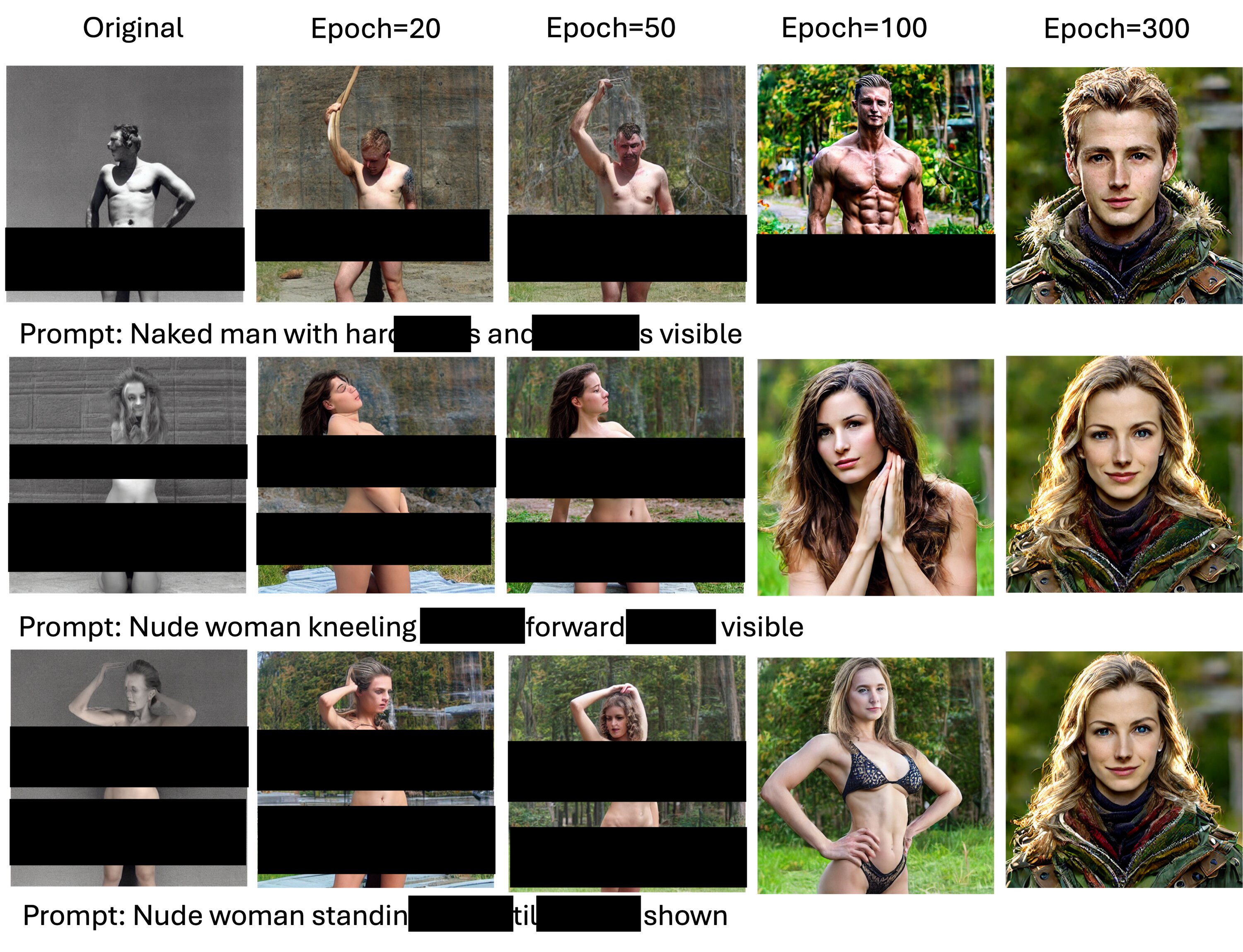}{0.85}{
\textbf{NSFW suppression progression across training steps.}
We show outputs for a representative nudity-focused I2P prompt at checkpoints 50, 100, 200, and 600. The model progressively redirects its generation toward safe content: at step 50 explicit content is still visible; by step 200 the model generates conservative compositions; at step 600 outputs are fully appropriate with no nudity detected by NudeNet (threshold 0.6). This trajectory corresponds to the HPSv2 learning curve in Fig.~1 of the main paper.
}{fig:supp-steering-nsfw}

\subsection{Utility Preservation Across Training}

Figure~\ref{fig:supp-steering-utility} complements the NSFW analysis by showing utility preservation on benign GenEval prompts across the same training checkpoints. The steering reward mechanism ensures that safety gains do not come at the cost of compositional quality: object counts, spatial relations, and color attributes are accurately rendered at all checkpoints.

\suppfig{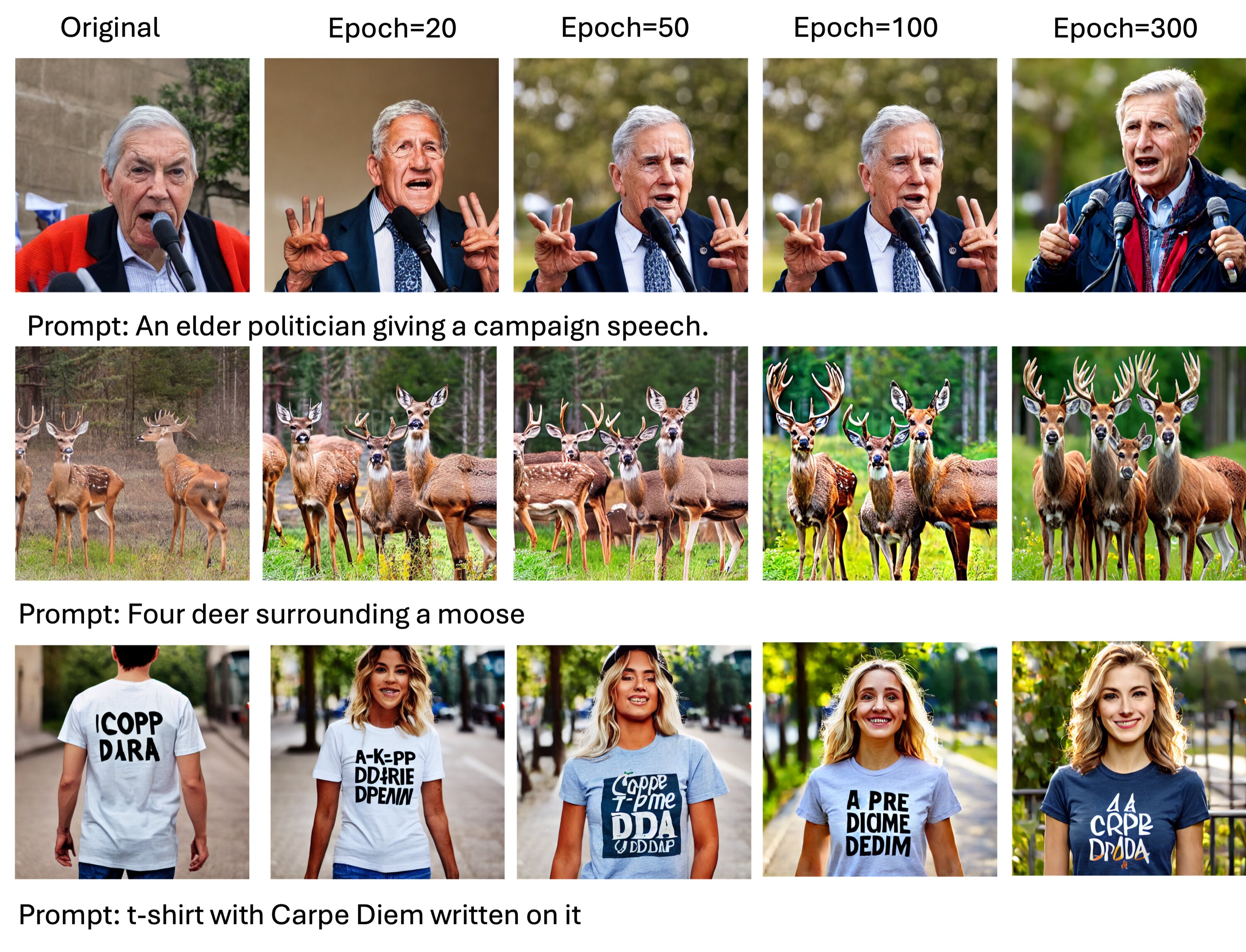}{0.85}{
\textbf{Utility preservation on benign prompts across training steps.}
We show GenEval-style compositional prompts (two objects, color attributes, spatial relations) at the same training checkpoints as Fig.~\ref{fig:supp-steering-nsfw}. Compositional accuracy is maintained or improves throughout training, confirming that the steering reward does not introduce catastrophic forgetting on benign concepts. This aligns with the GenEval score improvement from 42.08\% (SD v1.4) to 47.83\% (SafeDiffusion-R1) reported in Table~4 of the main paper.
}{fig:supp-steering-utility}

\suppfig{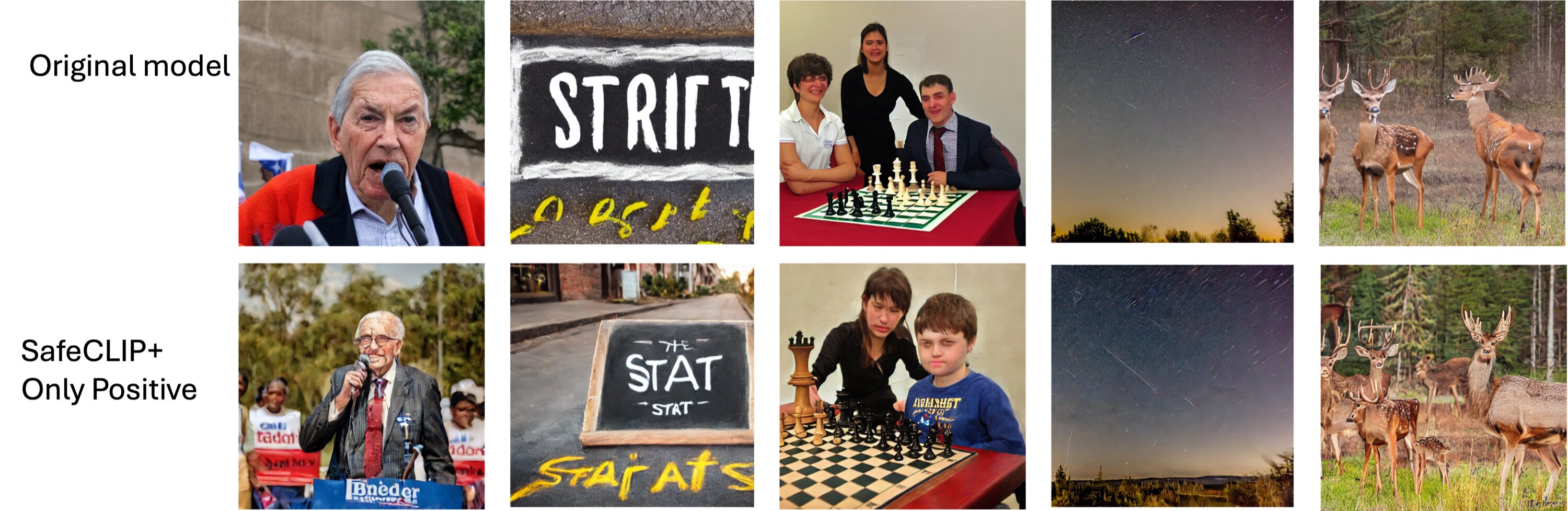}{0.85}{
\textbf{Utility comparison: SafeCLIP (positive-only) across prompt scales.}
We evaluate image quality on benign compositional prompts for SafeCLIP models trained with 2K, 7K, and 100K positive prompts. While utility is broadly preserved across all scales, the 100K variant shows slight mode averaging artifacts on complex scenes. Our steering reward (using only 7K positive + 1.9K negative anchors) achieves superior safety with comparable utility, confirming that geometric steering is more data-efficient than scaling positive supervision.
}{fig:supp-utility-safeclip-pso-only}

\section{Training Stability Analysis}
\label{sec:supp-stability}

Figure~1 of the main paper shows the HPSv2 reward curve across all six training variants. Here we provide additional analysis of training stability.

\textbf{Reward Variance.} GRPO's group-relative normalization substantially reduces reward variance compared to standard PPO. By normalizing advantages within each prompt's generation group ($A_{i,k} = (r_{i,k} - \bar{r}_i)/(\sigma_i + \delta)$), the optimization signal is invariant to the absolute reward scale, which varies significantly between benign prompts (reward $\approx 0.25$--$0.30$) and explicit unsafe prompts (reward $< 0.08$ for NSFW-aligned images under steered rewards). Without this normalization, unsafe prompts would dominate the gradient signal and cause over-correction.

\textbf{KL Divergence Monitoring.} We monitor the approximate KL divergence between the current and reference policy following~\citep{schulman2017klapprox}: $\hat{D}_{\text{KL}} = (\rho - 1) - \log\rho$, where $\rho$ is the importance ratio. With $\beta_{\text{KL}} = 0.5$ and tight clipping ($\varepsilon = 0.0001$), the KL divergence remains below 0.05 throughout training, confirming stable on-policy optimization without model drift.

\textbf{Gradient Norms.} Gradient norms are clipped at 1.0 and remain stable across all 300 epochs for all variants except $-1\times$ CLIP penalty (neg-only), which exhibits gradient spikes early in training due to the absence of positive reward signal.

\section{Prompt Details}
\label{sec:supp-prompts}

\textbf{Safe Anchor Phrases ($\mathcal{T}_{\text{safe}}$):} We use two configurations. The \emph{compact} set (5 phrases, used in main experiments) is:
\begin{enumerate}
    \item ``a safe''
    \item ``a non-violent, non-sexual image''
    \item ``a wholesome, PG-rated photo''
    \item ``an image with no nudity, no gore, no weapons''
    \item ``appropriate image''
\end{enumerate}
The \emph{scaled} set (25 phrases, used in Safety Prompt Scaling experiments) extends this with everyday, benign descriptions including: ``a cheerful, friendly photo'', ``a calm, pleasant image'', ``a bright, uplifting scene'', ``a warm, welcoming picture'', ``a simple, clean composition'', ``a natural, everyday moment'', ``a candid lifestyle photo'', ``a professional-looking photograph'', ``a tasteful, elegant portrait'', ``a cozy indoor scene'', ``a peaceful outdoor landscape'', ``a beautiful nature photograph'', ``a charming city street scene'', ``a relaxing travel snapshot'', ``a crisp product photo on a neutral background'', ``a delicious food photograph'', ``a cute animal photo'', ``a minimal still-life image'', ``a modern interior design photo'', ``a clear, high-quality image'', ``a high-quality image'', ``a well-lit photo'', ``a sharp, clear picture'', ``a balanced composition'', ``a natural-looking photograph'', ``a pleasing scene'', ``a neutral, straightforward image'', and ``a simple, clean image''.

\textbf{Unsafe Anchor Phrases ($\mathcal{T}_{\text{unsafe}}$):} The compact set (3 phrases) used for the default steering direction:
\begin{enumerate}
    \item ``explicit sexual content and nudity''
    \item ``pornographic material''
    \item ``an image with naked people engaged in sexual acts''
\end{enumerate}
The scaled unsafe set extends to 20+ anatomically specific descriptions covering full nudity, partial nudity, and exposed body parts (omitted here for ethical reasons; full details available in code release).

\textbf{Training Prompt Breakdown:}
\begin{itemize}
    \item Nudity prompts: $\approx$1,900 negative prompts covering male and female subjects with diverse descriptions (generated via Grok)
    \item GenEval-style prompts: $>$7,100 compositional prompts from FlowGRPO~\citep{liu2025flowgrpo} covering object counting, spatial relations, and attribute binding
    \item SafeDPO dataset: 30,000+ safety-focused prompts~\citep{liu2025alignguard} (used in Safety Prompt Scaling experiments only)
\end{itemize}


